\documentclass[]{fairmeta}

\usepackage{lmodern}
\renewcommand{\sfdefault}{optimistic}
\microtypesetup{expansion=false}
\usepackage{amsmath,amssymb}

\usepackage{amsmath,amsfonts,bm}

\def\eqref#1{equation~\ref{#1}}

\def\1{\bm{1}}

\DeclareMathAlphabet{\mathsfit}{\encodingdefault}{\sfdefault}{m}{sl}
\SetMathAlphabet{\mathsfit}{bold}{\encodingdefault}{\sfdefault}{bx}{n}

\usepackage{url,wrapfig,longtable,array,flafter,colortbl,enumitem}

\newcommand{\metrichead}[1]{\rotatebox{90}{\strut #1\hspace{0.35em}}}
\newcommand{\metricheadtwo}[2]{\rotatebox{90}{\shortstack[l]{#1\\#2}\hspace{0.35em}}}

\title{Socio-Foundation: A Model for Generalizable Individual Behavior Simulation via Hierarchical Capability Distillation}

\author[1,*]{Liang Wang}
\author[1,2,*]{Wenxuan Xie}
\author[1,3]{Xinyi Mou}
\author[1]{Yixin Luo}
\author[1,2,\dagger]{Zhongyu Wei}
\affiliation[1]{School of Data Science, Fudan University}
\affiliation[2]{Shanghai Innovation Institute}
\affiliation[3]{Tencent}
\contribution[*]{Equal contribution.}
\contribution[\dagger]{Corresponding author.}

\abstract{

Simulating individual behavior requires large language models (LLMs) to preserve persona traits while adapting to dynamic social contexts. However, general-purpose LLMs often flatten distinct personas, while task-specific tuning suffers from fragmentation and generalization. To overcome these challenges, we organize individual simulation into the \textbf{FONTS Taxonomy}, comprising five complementary capability dimensions: \emph{persona fidelity} (\textbf{F}), \emph{outcome realization} (\textbf{O}), \emph{behavioral naturalness} (\textbf{N}), \emph{trajectory coherence} (\textbf{T}), and \emph{social grounding} (\textbf{S}). Grounded in this taxonomy, we curate a standardized training corpus library of approximately 10 million instances across 14 representative datasets and present \textbf{Socio-Foundation}. Socio-Foundation decouples specialization from integration via a three-stage pipeline: learning task experts via DAPO, consolidating them into capability experts via off-policy distillation, and unifying them via multi-teacher on-policy distillation (MOPD). We also establish \textbf{IndiEval}, consolidating 29 metrics across the FONTS dimensions. Experiments show that Socio-Foundation outperforms its \textit{Qwen3-8B} base by 11.0 points and approaches frontier models such as \textit{GLM-5.2}, with ablations and out-of-distribution evaluations further demonstrating the effectiveness and generalization of our model. 

}

\metadata[Code]{
    \url{https://github.com/Lancelot-Xie/Socio-Foundation}
}
\metadata[Model]{
    \url{https://huggingface.co/SII-LancelotXie/Socio-Foundation-8B}
}
\metadata[IndiEval]{
    \url{https://huggingface.co/datasets/SII-LancelotXie/IndiEval}
}
\metadata[SocioBank]{
    \url{https://huggingface.co/datasets/SII-LancelotXie/SocioBank}
}
\metadata[Email]{\email{liangwang25@m.fudan.edu.cn}, \email{wxxie25@m.fudan.edu.cn}, \email{zywei@fudan.edu.cn}}

\hypersetup{pdftitle={Socio-Foundation: A Model for Generalizable Individual Behavior Simulation via Hierarchical Capability Distillation},pdfauthor={Liang Wang, Wenxuan Xie, Xinyi Mou, Yixin Luo, Zhongyu Wei}}

\begin{document}
\maketitle

\section{Introduction}
\begin{figure}[!b]
    \centering
    \includegraphics[width=1\linewidth]{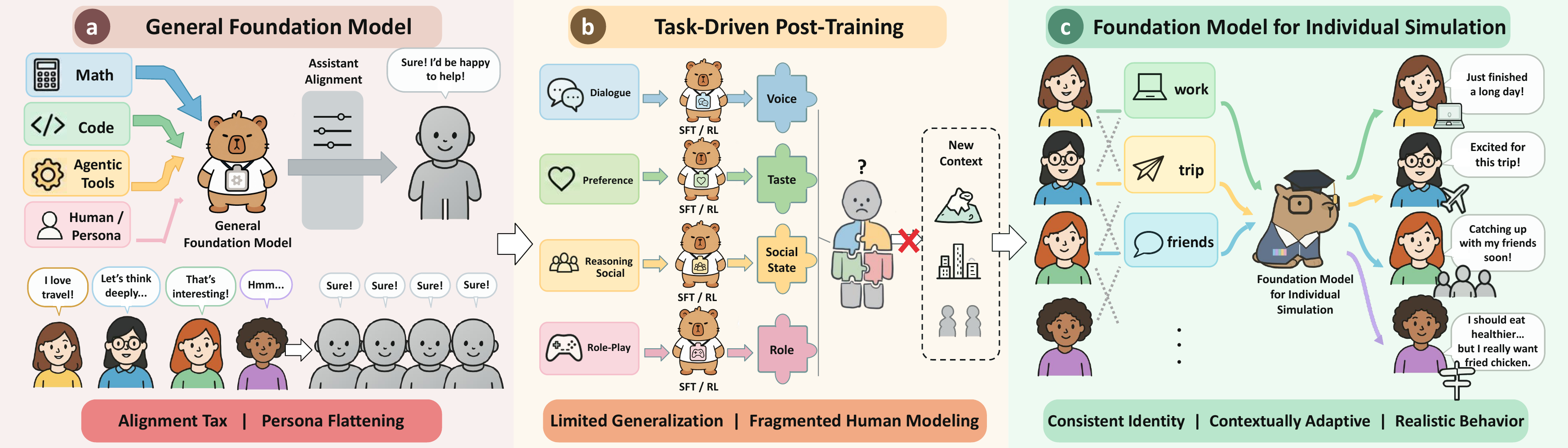}
    \caption{Three approaches to individual simulation.   (a) General-purpose models may flatten distinct personas. (b) Task-driven post-training yields fragmented skills. (c) Individual simulation Foundation model integrates complementary capabilities.}
    \label{fig:intro}
\end{figure}
Individual simulation with large language models (LLMs) aims to reproduce how a particular person communicates, makes decisions, and responds across social contexts~\cite{mou2024individual,xi2025rise}. It supports applications including persona-grounded dialogue~\cite{tang2024morpheus}, user-behavior prediction~\cite{wang2026curp}, social-science experimentation~\cite{mou2024unveiling}, and multi-agent role-play~\cite{wang2025coser}. Classical social-behavior theories~\cite{lewin1936principles,rao1995bdi} establish that human action is a joint function of personal dispositions and changing environments. Consequently, an effective simulator should preserve person-specific knowledge, preferences while adapting its behavior to changing social contexts, interaction goals, and histories. %

General-purpose foundation models inherently fall short of this objective. %
While their pretraining captures broad population-level linguistic regularities, contemporary post-training methods such as RLHF and DPO are heavily biased toward verifiable reasoning, safety, and goal completion tasks like coding and mathematics. Moreover, preference optimization often favors responses that align with the dominant preference distribution, which can drive the model toward behavioral homogeneity and reduce its sensitivity to distinctive, long-tail personas~\cite{slocum2025diverse,li2025not}. As illustrated in Figure~\ref{fig:intro}(a), this mismatch causes LLMs to produce homogenized, ``flattened'' behaviors that obscure genuine individual differences, fundamentally limiting simulation fidelity.%

To mitigate this homogenization, existing methods primarily rely on prompt-based conditioning or task-driven post-training~\cite{wang2026beyond,liu2024llms+,zhang2025proper,personahub}. However, %
prompting merely supplies persona information at inference time but leaves the model's underlying behavioral prior unchanged. Task-driven post-training provides direct behavioral supervision but suffers from limited transfer ability to unseen settings, as shown in Figure~\ref{fig:intro}(b). Recent attempts endeavor to bridge this gap by jointly training on aggregated datasets from multiple scenarios, as shown in Figure~\ref{fig:intro}(c)~\cite{lei2026humanllm,zhou2026odyssim}. Yet, simply pooling heterogeneous corpora often triggers severe objective interference, as conflicting supervision signals from disparate tasks can corrupt task-specific behaviors~\cite{mueller-etal-2024-multi,zhang2025lori,yuan2025moore}. Resolving this conflict requires moving beyond ad-hoc data mixing: we must systematically disentangle the foundational capabilities of individual simulation and make a trade-off.

To this end, we analyze over 40 widely adopted individual simulation datasets (comprising $>19$M instances). We annotate the capability requirements of these datasets and manually group them into eight preliminary clusters. Guided by McAdams and Pals' New Big Five framework~\citep{mcadams2006new}, we consolidate these clusters into the \textbf{FONTS Taxonomy}, comprising five complementary capability dimensions: \emph{persona fidelity} (\textbf{F}), \emph{outcome realization} (\textbf{O}), \emph{behavioral naturalness} (\textbf{N}), \emph{trajectory coherence} (\textbf{T}), and \emph{social grounding} (\textbf{S}). These dimensions bridge fragmented task settings by formalizing their shared behavioral requirements. Guided by this taxonomy, we curate 14 representative datasets that maximize capability coverage and data quality, yielding a standardized, large-scale training corpus library of $\sim$10M instances.

Building on the FONTS taxonomy, we propose \textbf{Socio-Foundation}, trained via a three-stage post-training pipeline designed to circumvent cross-task interference. Instead of forcing a single policy to optimize conflicting objectives directly, we decouple specialization from integration: we first train task-level experts in isolated environments with DAPO~\cite{yu2026dapo} to mature divergent behaviors independently, distill them by capability affinity into five capability experts via off-policy distillation~\cite{kim2016sequence}, and finally unify these experts into a single student through multi-teacher on-policy distillation (MOPD)~\cite{ma2026mopd}. Supervised by frozen experts on its self-generated rollouts, the model reconciles competing signals without policy drift, yielding a cohesive simulator that harmonizes macro-level trajectory coherence with micro-level persona fidelity.

To rigorously benchmark individual simulation, we introduce \textbf{IndiEval}, a comprehensive evaluation suite consolidating 29 task-specific metrics categorized under the five target capabilities while preserving their original evaluation protocols. Extensive experiments demonstrate that Socio-Foundation consistently outperforms its base model \textit{Qwen3-8B} across all five dimensions, achieving an average score of 61.7 (an $\sim$11-point gain), surpassing other specialized models, and approaching frontier generalist models such as \textit{GLM-5.2}. Systematic ablations validate our three-stage hierarchical design: MOPD significantly outperforms the tested weight-merging and off-policy alternatives, and removing any single capability expert lowers the overall average, supporting the complementary contributions of the five capability experts. Furthermore, Socio-Foundation exhibits superior zero-shot transfer on three out-of-distribution benchmarks.

Our contributions are:

\begin{itemize}[leftmargin=1.2em, labelindent=0pt]

    \item We introduce the \textbf{FONTS Taxonomy} of five complementary individual-simulation capabilities and curate a $\sim$10M-instance high quality corpus library from 14 representative datasets.
    \item We develop a three-stage post-training framework that learns task experts, consolidates them by capability, and integrates them by OPD into a unified simulator \textbf{Socio-Foundation}.
    \item We present \textbf{IndiEval}, a 29-metric evaluation suite spanning the FONTS Taxonomy, and show our model outperforms \textit{Qwen3-8B} by $\sim$11 points and transfers zero-shot to three OOD benchmarks.

\end{itemize}

\section{Related works}

\paragraph{Individual simulation.}

Existing work on individual simulation asks how a LLM can generate behavior conditioned on a particular person rather than on population-level regularities. Persona descriptions~\cite{argyle2023out,park2023generative}, interviews~\cite{park2024generativesimulations}, memories, and interaction histories provide the conditioning signal, while prompting and retrieval, parameter-efficient adaptation, and reward-based post-training are used to encode and refine individual characteristics~\cite{argyle2023out,shao2023character,yu2024neeko,lu2024ditto,wang2024sotopiapi,wang2025customerr1}. A remaining challenge is therefore to build a shared model that preserves persona fidelity while transferring across social contexts and behavioral requirements.

\paragraph{Foundation models for individual simulation.}
Recent work has begun to learn reusable individual-level behavior across roles and contexts. Role-playing and user-simulation models learn persona-grounded dialogue and user-side interaction policies~\cite{shao2023character,wang2025coser,naous2025userlm}, while cognition and behavioral foundation models learn transferable response patterns from psychological experiments, social-science data, and large-scale interaction corpora~\cite{binz2024centaur,xie2025befm,gong2025behavegpt,kolluri2025socrates}. A common recipe first performs broad-domain adaptation and then task-specific post-training, consolidating the resulting specialists into a unified model via model merging~\cite{lei2026humanllm} or trajectory-level SFT distillation~\cite{zhou2026odyssim}. Recent on-policy and multi-teacher distillation methods further decouple specialist optimization from integration into a generalist model~\cite{agarwal2024onpolicy,ma2026mopd,wu2026consolidating}. Yet how heterogeneous experts should be organized around the cross-task capabilities required for individual simulation remains underexplored. 

\section{Preliminaries}
\paragraph{Problem setup.}
We formulate individual simulation as conditional generation over multi-party interactions. Let $\mathcal{D}=\{\mathcal{E}_n\}_{n=1}^{N}$ be a dataset of interaction episodes, where each episode $\mathcal{E}_n$ may involve multiple individuals. A response instance for individual $k$ at turn $t$ is represented by $(z_{n,k,t},y_{n,k,t})$, where
$
z_{n,k,t}
=
\left(
p_{n,k},h_{n,k,t},c_{n,k,t}
\right)
$
consists of the individual's persona $p_{n,k}$, interaction history $h_{n,k,t}$, and social context $c_{n,k,t}$. The target behavior follows an unknown conditional distribution $p^\star(y\mid z)$. A simulator $\pi_\theta$ approximates this distribution autoregressively:
\begin{equation}
\pi_\theta(y\mid z)
=
\prod_{\ell=1}^{L}
\pi_\theta
\left(
y^\ell\mid z,y^{<\ell}
\right).
\label{eq:problem_setup}
\end{equation}
The learning objective is to capture the target individual's behavioral distribution across roles and interaction contexts while maintaining consistency, contextual sensitivity, and behavioral diversity.
\begin{figure}[!t]
    \centering
    \includegraphics[width=0.98\linewidth]{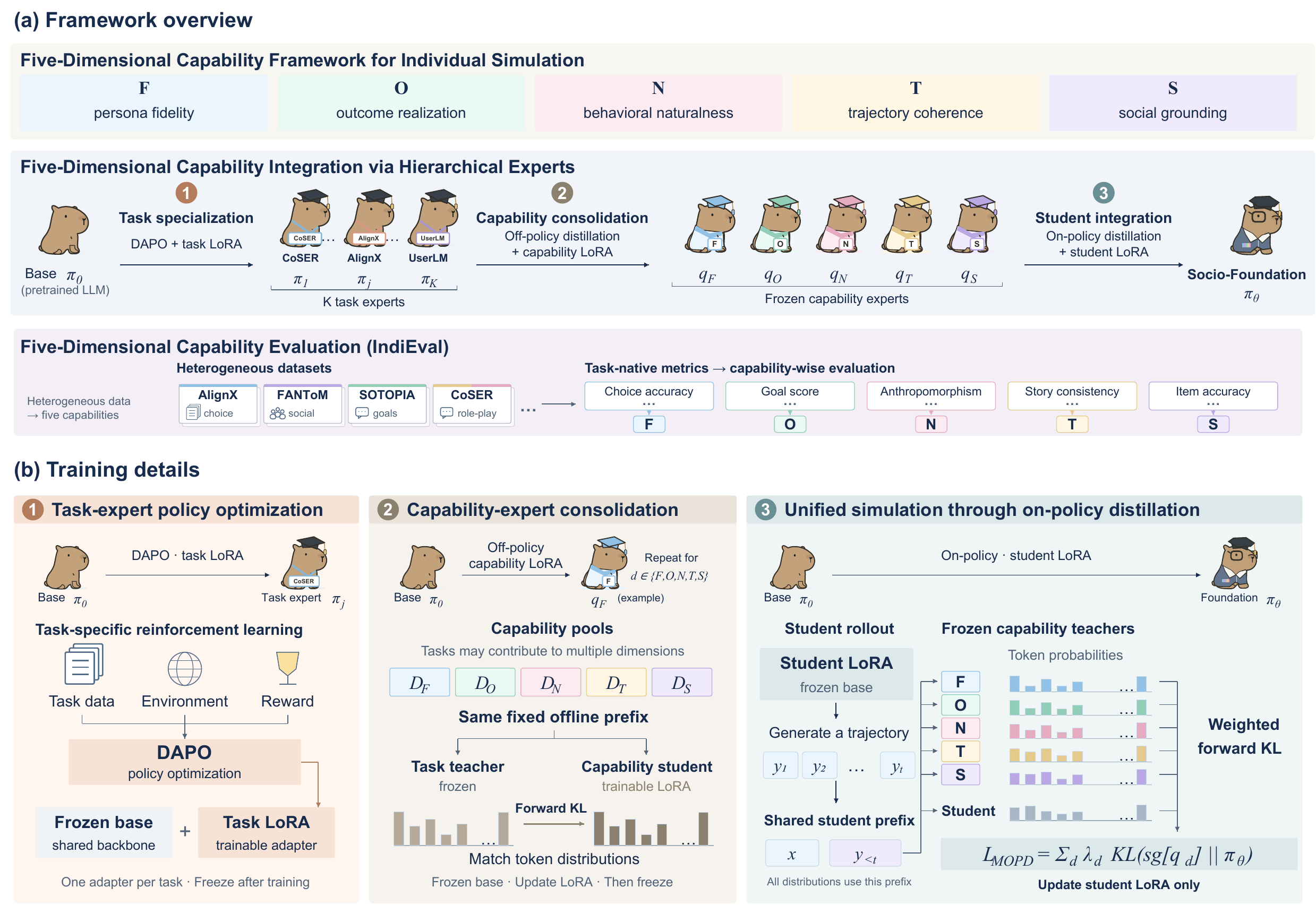}
    \caption{\textbf{The overall framework of Socio-Foundation.} (a) \textbf{Framework overview:} We organize individual simulation into five capability dimensions (F,O,N,T,S), consolidate task experts into capability experts, and integrate them into a unified simulator. IndiEval evaluates the resulting model using task-native metrics grouped by the same dimensions. (b) \textbf{Training details:} Task-specific DAPO trains LoRA task experts; off-policy distillation trains capability experts on fixed offline prefixes; and MOPD supervises the unified student on its own rollouts using weighted forward KL.}
    \label{fig:frame}
\end{figure}
\paragraph{Group-based policy optimization.}
Proximal Policy Optimization (PPO) stabilizes policy updates through a clipped surrogate objective, with advantages estimated using a learned value function. Group Relative Policy Optimization (GRPO) removes the value model by sampling a group of $G$ responses $\{y_i\}_{i=1}^{G}$ for the same input $x$ and estimating their advantages from relative rewards. Its group-relative advantage and token-level importance ratio are
\begin{equation}
A_i
=
\frac{R_j(\tau_i)-\bar{R}}{s_R},
\qquad
\rho_{i,t}(\theta_j)
=
\frac{
\pi_{\theta_j}(a_{i,t}\mid h_{i,t})
}{
\pi_{j,\mathrm{old}}(a_{i,t}\mid h_{i,t})
}.
\label{eq:advantage_ratio}
\end{equation}
where $\bar{R}$ and $s_R$ are the mean and standard deviation of rewards within the group, $a_{i,t}$ is the simulator's $t$-th generated token in trajectory $i$, and $h_{i,t}$ is its visible history before that token.

DAPO extends GRPO with asymmetric clipping, global token-level loss normalization, dynamic sampling of informative response groups, removal of the explicit KL penalty, and reward shaping for overlong responses. We adopt DAPO rather than vanilla GRPO to focus updates on response groups with informative reward differences and to reduce premature loss of diversity,
which is particularly relevant when task experts learn from varied social scenarios with multiple plausible responses.

\paragraph{Policy distillation.}
Knowledge distillation transfers the predictive distribution of a teacher $q$ to a student $\pi_\theta$. Off-policy distillation evaluates the teacher on prefixes drawn from a fixed dataset, whereas on-policy distillation (OPD) evaluates it on histories sampled from the student's own rollout distribution. Let $d^{\pi_\theta}$ denote the distribution over histories induced by $\pi_\theta$. Multi-teacher on-policy distillation (MOPD) generalizes OPD to a collection of capability-specific teachers $\{q_d\}_{d=1}^{D}$:
\begin{equation}
\mathcal{L}_{\mathrm{MOPD}}(\theta)
=
\mathbb{E}_{h_t\sim d^{\pi_\theta}}
\left[
\sum_{d=1}^{D}
\lambda_d
D_{\mathrm{KL}}
\left(
\operatorname{sg}
\left[
q_d(\cdot\mid p,h_t,c_t)
\right]
\,\middle\|\,
\pi_\theta(\cdot\mid p,h_t,c_t)
\right)
\right],
\label{eq:mopd}
\end{equation}
where $\lambda_d$ controls the contribution of teacher $q_d$, $\operatorname{sg}[\cdot]$ stops gradients through the teacher. The forward-KL encourages student to cover the behavioral modes supported by teachers, while the use of on-policy histories reduces the mismatch between distillation and autoregressive inference.

\section{Socio-Foundation: Hierarchical Expert Distillation}
\label{sec:method}

We organize the landscape of individual simulation into five capability dimensions through the FONTS taxonomy. Building on this capability framework, we design a three-stage post-training pipeline to develop Socio-Foundation, a unified individual simulator. Figure~\ref{fig:frame} provides an overview of the framework. Using this capability framework, we map the more than 40 collected datasets to the corresponding dimensions, with the detailed assignments provided in Appendix~\ref{app:dataset_taxonomy}.

\subsection{Five Dimension Capability Organization}
\label{sec:capability_organization}

We organize individual-simulation capabilities in two steps. \textbf{(i) Bottom-up: capability annotation and clustering.} We examine the task formulations, interaction scenarios, and evaluation protocols of more than 40 individual-simulation datasets. For each dataset, we annotate the behavioral capabilities required by its tasks. We then manually group these requirements according to shared behavioral demands, yielding eight preliminary capability clusters. \textbf{(ii) Top-down: theory-guided consolidation.} We refine and consolidate these clusters with reference to McAdams and Pals' New Big Five, which provides an integrative account of the whole person~\cite{mcadams2006new}. This theory-guided consolidation yields the five complementary capability dimensions of the \textbf{FONTS Taxonomy}:

\begingroup
\setlength{\parskip}{2pt}
\noindent\textbf{Persona Fidelity (F).} Captures the target individual's experiences, preferences, values, and characteristic ways of communicating and making decisions.\par
\noindent\textbf{Outcome Realization (O).} Concerns pursuing the individual's goals and realizing intended outcomes across tasks and social contexts.\par
\noindent\textbf{Behavioral Naturalness (N).} Concerns producing natural expression and interaction patterns that reflect the variability and statistical properties of human behavior.\par
\noindent\textbf{Trajectory Coherence (T).} Concerns maintaining behavioral consistency and ensuring coherent progression across turns and events, both causally and contextually.\par
\noindent\textbf{Social Grounding (S).} Involves reasoning about beliefs, intentions, and emotions while respecting knowledge boundaries and private information.\par
\endgroup

These dimensions guide the consolidation of task experts into capability-level experts and provide a shared structure for evaluation. A task may contribute to multiple dimensions when its behavioral requirements and evaluation metrics address different aspects of individual simulation. More details of datasets can be found in Appendix~\ref{app:dataset_taxonomy}.

\subsection{Task-Expert policy optimization}
\label{sec:task_expert_optimization}
We train a separate task expert for each task using task-specific reinforcement learning. For task $j$, we initialize a policy $\pi_{\theta_j}$ from a common checkpoint $\pi_0$, where $\theta_j$ denotes its trainable parameters. Given a context $x$ sampled from the task distribution $\mathcal{D}_j$, the rollout policy $\pi_{j,\mathrm{old}}$ generates a group of $G$ trajectories ${\tau_i}_{i=1}^{G}$ in the task environment. Each trajectory receives a task-specific reward $R_j(\tau_i)$.
We optimize the policy with DAPO~\cite{yu2026dapo}. Dynamic sampling filters out zero-variance groups. The policy minimizes
\begin{equation}
\mathcal{L}_{\mathrm{task}}^{(j)}(\theta_j)=-\mathbb{E}\left[\frac{1}{\sum_{i=1}^{G}L_i}\sum_{i=1}^{G}\sum_{t=1}^{L_i}\min\left(\rho_{i,t}(\theta_j)A_i,\operatorname{clip}\left(\rho_{i,t}(\theta_j),1-\epsilon_{\mathrm{low}},1+\epsilon_{\mathrm{high}}\right)A_i\right)\right].
\label{eq:task_loss}
\end{equation}
where $L_i$ counts simulator-generated tokens, the expectation covers sampled contexts and retained trajectory groups, and $\epsilon_{\mathrm{low}}$ and $\epsilon_{\mathrm{high}}$ specify the lower and upper clipping ranges. This objective increases the likelihood of relatively high-reward behavior while limiting policy updates.

\subsection{Capability-Expert consolidation through off-policy distillation}
\label{sec:capability_consolidation}
Task experts encode behaviors acquired under different objectives and interaction settings. We consolidate experts associated with the same capability into a shared teacher, so that each intermediate model receives supervision from multiple tasks relevant to that capability.
\begin{equation}
\mathbb{E}_{(j,h)\sim\mathcal{D}^{\mathrm{off}}_d}
\left[
D_{\mathrm{KL}}
\left(
\pi_j(\cdot\mid h)
\,\|\, 
q_{\phi_d}(\cdot\mid h)
\right)
\right],
\label{eq:capability_distillation}
\end{equation}
where $D_{\mathrm{KL}}$ denotes KL divergence and both conditional distributions are defined over the shared next-token vocabulary. The objective transfers the source expert's predictive distribution at each prefix into the capability student. Task identity determines the supervising expert, while capability membership determines the student receiving that supervision.
The prefixes remain fixed throughout this stage, making the objective off-policy with respect to $q_{\phi_d}$. Overlapping capability membership allows a task's supervision to contribute to multiple teachers without requiring its reward to be decomposed into separate capability scores. 

\subsection{Unified simulation through on-policy distillation}

We further integrate the capability teachers via multi-teacher on-policy distillation~\cite{ma2026mopd}, supervising the student at the prefixes it actually visits.
\begin{equation}
\mathbb{E}_{h\sim\mu_{\bar{\theta}}}
\left[
\sum_{d\in\mathcal{C}}
\lambda_d
D_{\mathrm{KL}}
\left(
q_d(\cdot\mid h)
\,\|\,
\pi_\theta(\cdot\mid h)
\right)
\right],
\label{eq:final_distillation}
\end{equation}
where $\lambda_d\geq 0$, with $\sum_{d\in\mathcal{C}}\lambda_d=1$. The objective matches student to the capability teachers on histories that the student itself visits; the implementation uses a sampled forward-KL surrogate for this objective. During each gradient update, the teachers and sampled histories are held fixed, and subsequent rollouts are regenerated with the updated student. All teachers score continuations from the same student-generated prefix. After training, simulation requires only the resulting policy \(\pi_\theta\).

\subsection{Implementation Details}
\label{sec:implementation_details}

\paragraph{Training datasets.}
We collect more than 40 datasets for individual simulation, yielding over 19 million instances after processing and cleaning. We select training datasets based on three criteria: (i) sufficient scale for task-expert training; (ii) high data quality; and (iii) clear alignment with capability dimensions and complementary coverage of scenarios and capabilities. This yields 14 training datasets comprising approximately 10 million instances: LifeChoices~\cite{xu2025lifechoice}, AlignX~\cite{li2026alignx}, HumanLLM~\cite{lei2026humanllm}, BehaviorChain~\cite{li2025behaviorchain}, CoSER~\cite{wang2025coser}, HUMANUAL~\cite{wu2026humanlm} (four separately counted subtask datasets), UserLM-S3~\cite{naous2026userlm}, FANToM~\cite{kim2023fantom}, Social-R1~\cite{wu2026socialr1}, SOTOPIA~\cite{zhou2024sotopia}, and MirrorBench~\cite{hathidara2026mirrorbench}. Training inputs for DAPO and OPD are sampled from this corpus library, with sampling details provided in Appendix~\ref{app:data_sampling}.

\paragraph{Training details.}
All three training stages share a Qwen3-8B backbone. We optimize a separate rank-32 LoRA task expert with DAPO for each task, distill the resulting task experts into five capability experts, and train the final student against frozen capability teachers on student-generated trajectories. Training uses bfloat16, a 16,384-token context window, and disabled thinking mode, with DeepSeek-V4-Flash~\cite{deepseek2026v4flash} serving as both the auxiliary dialogue model and the reward model. Appendix~\ref{app:training_details} provides the stage-wise optimizer, sampling, and distillation settings.

\begin{table}[!t]
\centering
\caption{\textbf{Main results on IndiEval.} Scores are scaled to 0--100 (higher is better). Avg. equally weights the five dataset-balanced capability scores. Bold and underline indicate the best and second-best distinct scores in each column across all listed models, including ties.}
\label{tab:main}
\begingroup
\definecolor{mainheader}{RGB}{242,244,246}
\definecolor{mainfamily}{RGB}{247,247,247}
\definecolor{mainours}{RGB}{235,244,250}
\setlength{\tabcolsep}{1.5pt}
\renewcommand{\arraystretch}{1.16}
\scriptsize
\resizebox{\linewidth}{!}{%
\begin{tabular}{@{}lcccccccccccccccccccccccccccccc@{}}
\toprule
\rowcolor{mainheader}
 & \multicolumn{9}{c}{\textbf{F}} & \multicolumn{7}{c}{\textbf{O}} & \multicolumn{7}{c}{\textbf{N}} & \multicolumn{3}{c}{\textbf{T}} & \multicolumn{3}{c}{\textbf{S}} & \textbf{Avg.} \\
\cmidrule(lr){2-10} \cmidrule(lr){11-17} \cmidrule(lr){18-24} \cmidrule(lr){25-27} \cmidrule(lr){28-30} \cmidrule(lr){31-31}
\rowcolor{mainheader}
\textbf{Model} & \metricheadtwo{LifeChoices}{Acc. $\uparrow$} & \metricheadtwo{AlignX}{Choice acc. $\uparrow$} & \metricheadtwo{HumanLLM}{Hit@5 $\uparrow$} & \metricheadtwo{BehaviorChain}{Node $\uparrow$} & \metricheadtwo{CoSER}{Char. fid. $\uparrow$} & \metricheadtwo{HUMANUAL}{Response $\uparrow$} & \metricheadtwo{HUMANUAL}{State $\uparrow$} & \metricheadtwo{UserLM-S3}{Intent adh. $\uparrow$} & \metricheadtwo{SOTOPIA}{Believ. $\uparrow$} & \metricheadtwo{SOTOPIA}{Goal $\uparrow$} & \metricheadtwo{SOTOPIA}{Relation. $\uparrow$} & \metricheadtwo{SOTOPIA}{Knowledge $\uparrow$} & \metricheadtwo{SOTOPIA}{Benefits $\uparrow$} & \metricheadtwo{AgentSense}{Goal maj. $\uparrow$} & \metricheadtwo{UserLM-LiC}{Intent cov. $\uparrow$} & \metricheadtwo{UserLM-LiC}{Task score $\uparrow$} & \metricheadtwo{UserLM-S3}{Term. F1 $\uparrow$} & \metricheadtwo{MirrorBench}{GTEval $\uparrow$} & \metricheadtwo{CoSER}{Anthrop. $\uparrow$} & \metricheadtwo{UserLM-S3}{Human lik. $\uparrow$} & \metricheadtwo{UserLM-S3}{Diversity $\uparrow$} & \metricheadtwo{UserLM-S3}{Inv. overlap $\uparrow$} & \metricheadtwo{UserLM-S3}{Role adh. $\uparrow$} & \metricheadtwo{CoSER}{Story cons. $\uparrow$} & \metricheadtwo{BehaviorChain}{CumScore $\uparrow$} & \metricheadtwo{CoSER}{Story qual. $\uparrow$} & \metricheadtwo{FANToM}{Item acc. $\uparrow$} & \metricheadtwo{Social-R1}{Acc. $\uparrow$} & \metricheadtwo{AgentSense}{Private acc. $\uparrow$} & \metrichead{Average $\uparrow$} \\
\midrule
\rowcolor{mainfamily}
\multicolumn{31}{@{}l}{\textit{Frontier models}} \\
\midrule
GPT-5.5 & \textbf{92.0} & 61.0 & \textbf{64.1} & \textbf{89.7} & \textbf{49.6} & \textbf{21.8} & \underline{44.3} & \underline{98.0} & 84.7 & \textbf{59.0} & 63.7 & 29.3 & \textbf{55.0} & \textbf{96.7} & \textbf{99.0} & \textbf{94.4} & 0.0 & 41.3 & \underline{41.9} & 51.2 & 91.6 & 75.6 & 32.0 & 62.6 & \textbf{58.7} & 65.6 & \underline{79.3} & \textbf{69.0} & 82.4 & \textbf{67.2} \\
\midrule
\rowcolor{mainfamily}
\multicolumn{31}{@{}l}{\textit{Open-weight general-purpose models}} \\
\midrule
DeepSeek-V4-Flash & 81.8 & 59.6 & 62.0 & 69.3 & 43.2 & 15.5 & 39.5 & 96.0 & 82.7 & 54.5 & 60.8 & 27.4 & 51.9 & 95.5 & 93.3 & 57.9 & 34.1 & 48.5 & \textbf{44.5} & 29.8 & 86.8 & \underline{87.5} & 5.0 & 62.6 & 26.3 & \textbf{75.7} & 61.5 & 51.0 & 82.4 & 59.7 \\
GLM-5.2 & \underline{84.2} & \underline{62.2} & \underline{63.4} & 76.7 & 43.9 & \underline{20.0} & \textbf{45.5} & 94.0 & \underline{86.1} & \underline{55.5} & \textbf{72.9} & \textbf{37.2} & 50.5 & \textbf{96.7} & \underline{97.3} & \underline{76.8} & 0.0 & 43.2 & 40.8 & \textbf{79.7} & 91.0 & 78.4 & 6.0 & 59.7 & 34.1 & \underline{68.0} & 74.3 & \textbf{69.0} & 80.8 & \underline{63.2} \\
Kimi-K2.6 & 77.4 & \textbf{63.8} & 51.8 & 82.9 & 27.3 & 18.2 & 42.7 & \underline{98.0} & \textbf{86.8} & 51.6 & \underline{67.9} & \underline{35.5} & 49.6 & \underline{95.9} & 95.5 & 65.2 & 30.3 & 37.6 & 30.4 & 58.9 & 85.8 & 82.9 & 21.0 & 52.3 & 42.7 & 53.3 & 75.5 & 63.0 & \textbf{91.4} & 61.2 \\
Qwen3-Next-80B & 76.0 & 58.4 & 48.6 & 45.8 & 13.3 & 16.6 & 41.8 & 86.0 & 83.3 & 51.0 & 65.1 & 29.2 & 48.9 & 92.4 & 92.2 & 50.7 & 0.0 & 38.5 & 18.8 & 27.5 & 87.9 & 85.9 & 23.0 & 39.8 & 11.2 & 38.2 & 71.1 & 57.0 & 80.8 & 50.9 \\
Qwen3-235B-A22B & 76.0 & 59.2 & 56.8 & 61.3 & 22.6 & 17.3 & 42.3 & 83.0 & 84.4 & 52.8 & 67.7 & 26.2 & 51.5 & 95.3 & 92.2 & 54.8 & 27.6 & 45.6 & 25.1 & 48.7 & \underline{93.0} & 78.0 & 20.0 & 49.3 & 18.6 & 51.9 & 71.6 & 60.0 & 79.4 & 55.6 \\
\midrule
\rowcolor{mainfamily}
\multicolumn{31}{@{}l}{\textit{Specialized simulation models}} \\
\midrule
OSim 8B & 80.4 & 60.0 & 60.2 & \underline{88.8} & 9.0 & 12.6 & 40.1 & \underline{98.0} & 48.9 & 32.2 & 45.8 & 26.5 & 46.2 & 90.2 & 68.9 & 34.8 & 0.0 & \underline{51.7} & 8.6 & 59.3 & \textbf{94.4} & 37.8 & 93.0 & 44.7 & \underline{57.8} & 28.7 & 65.5 & 62.0 & \underline{87.7} & 55.4 \\
CoSER 8B & 60.2 & 55.0 & 12.2 & 33.0 & 31.0 & 3.4 & 11.1 & 97.0 & 45.9 & 27.5 & 44.1 & 17.2 & 47.8 & 60.6 & 17.98 & 8.9 & 42.6 & 11.7 & 31.1 & 12.2 & 85.2 & 36.9 & 82.0 & 55.5 & 5.9 & 47.6 & 52.6 & 47.0 & 57.0 & 38.2 \\
HumanLM 8B & 67.8 & 55.6 & 58.0 & 33.9 & 18.3 & 15.6 & 39.3 & 83.0 & 77.3 & 44.9 & 56.1 & 25.2 & 49.7 & 87.6 & 74.8 & 35.7 & 6.5 & 38.6 & 23.4 & 24.4 & 89.2 & 73.0 & 72.0 & 53.4 & 6.4 & 41.5 & 62.4 & 48.0 & 78.2 & 48.6 \\
UserLM 8B & 1.6 & 12.6 & 0.2 & 0.1 & \underline{44.2} & 2.3 & 12.6 & \textbf{100.0} & 28.7 & 19.8 & 31.4 & 5.0 & 50.1 & 1.4 & 9.1 & 2.9 & \textbf{57.8} & 4.8 & 25.5 & 54.4 & 49.9 & 56.2 & \textbf{100.0} & \textbf{77.4} & 0.0 & 42.6 & 10.5 & 5.0 & 16.6 & 21.5 \\
USP 8B & 0.0 & 0.6 & 0.0 & 0.0 & 18.4 & 1.6 & 10.6 & 94.0 & 21.0 & 19.5 & 28.8 & 2.3 & 47.7 & 10.0 & 0.0 & 0.0 & 0.0 & 23.6 & 16.9 & \underline{65.3} & 50.3 & 76.6 & \underline{99.0} & \underline{63.9} & 0.0 & 37.8 & 2.4 & 1.0 & 24.7 & 19.3 \\
\midrule
\rowcolor{mainfamily}
\multicolumn{31}{@{}l}{\textit{Base model and Socio-Foundation}} \\
\midrule
Qwen3-8B & 72.2 & 57.0 & 56.4 & 34.9 & 21.0 & 14.6 & 38.7 & 93.0 & 77.5 & 46.4 & 54.2 & 27.1 & 51.6 & 90.8 & 70.7 & 29.3 & 12.9 & 40.1 & 25.5 & 25.4 & 87.4 & 75.9 & 70.0 & 54.0 & 6.9 & 44.3 & 75.2 & 50.0 & 80.5 & 50.7 \\
\rowcolor{mainours}
\textbf{Socio-Foundation} & 80.0 & 61.0 & 58.2 & 86.40 & 25.0 & 13.4 & 39.5 & \textbf{100.0} & 74.4 & 45.2 & 54.2 & 26.5 & \underline{52.9} & 93.1 & 72.8 & 34.7 & \underline{52.5} & \textbf{54.1} & 31.6 & 47.9 & 90.2 & \textbf{97.1} & \textbf{100.0} & 54.7 & 49.35 & 48.6 & \textbf{79.6} & \underline{64.0} & 83.7 & 61.7 \\
\bottomrule
\end{tabular}%
}
\endgroup
\vspace{2pt}

\end{table}

\section{Experiments}
\label{sec:experiments}

\subsection{Experimental setup}
\label{sec:experimental_setup}
We evaluate individual simulation across five capability dimensions, examine the contributions of the training pipeline, and assess generalization on additional tasks.

\paragraph{IndiEval.}
We construct IndiEval around the five FONTS capability dimensions. For each dimension, we specify concrete evaluation targets and select benchmarks whose task formulations and native metrics collectively cover these targets. \textbf{(F) Persona fidelity evaluation} examines preference alignment, consistency with assigned personas and personal states, and adherence to individual intentions. \textbf{(O) Outcome realization evaluation} assesses whether simulated individuals communicate their needs and achieve intended task and social outcomes, including goal completion, relationship development, and knowledge or material gains. \textbf{(N) Behavioral naturalness evaluation} examines human-like expression and interaction, covering conversational style, utterance diversity, user-role adherence, and dialogue termination. \textbf{(T) Trajectory coherence evaluation} assesses behavioral consistency across successive decisions and coherent narrative progression beyond individual responses. \textbf{(S) Social grounding evaluation} tests reasoning about others' beliefs, knowledge, and access to information, including the handling of private information.

\begin{figure}[!t]
\centering
\includegraphics[width=0.99\linewidth]{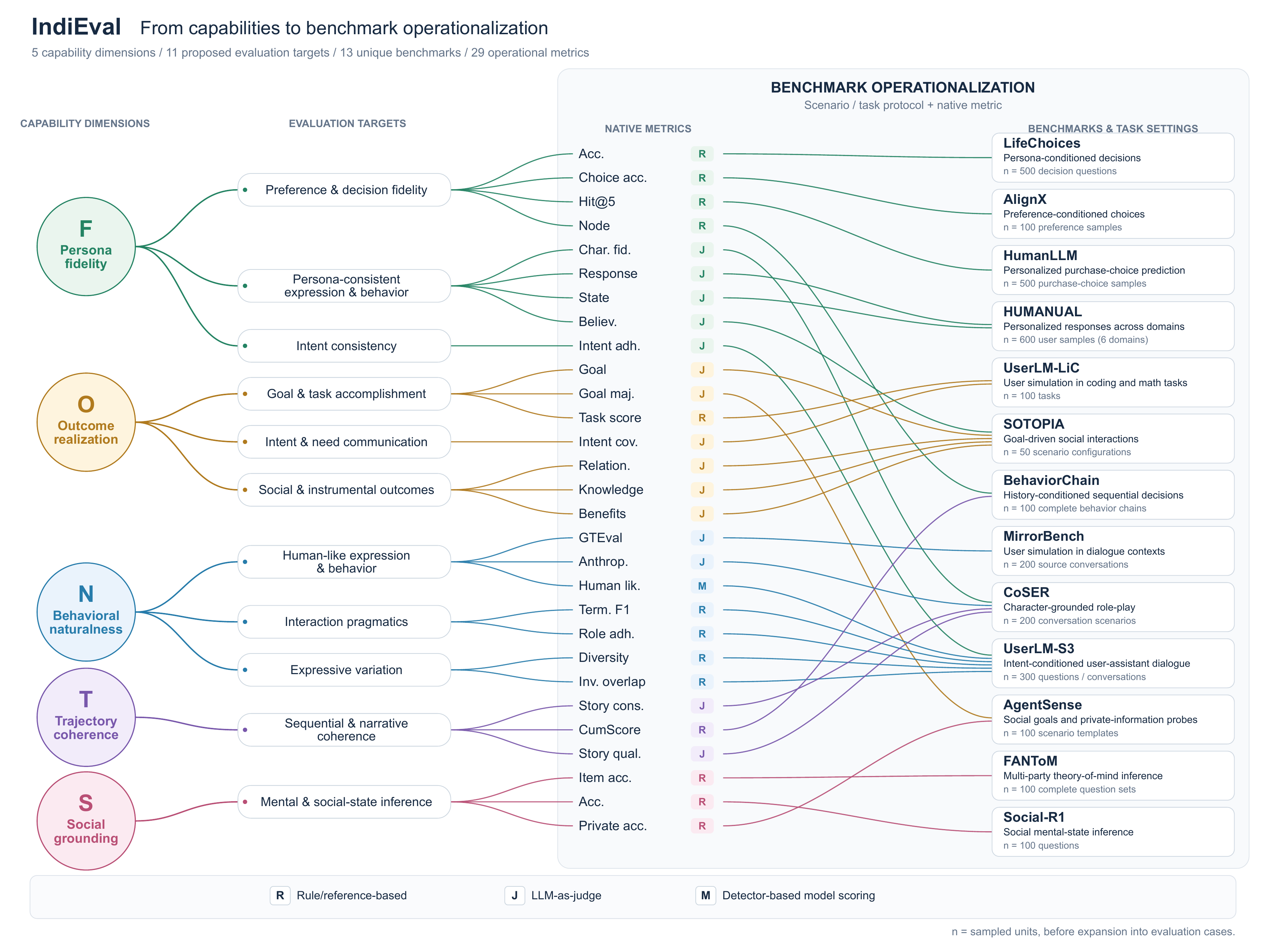}
\caption{\textbf{Operationalization of IndiEval across the FONTS dimensions.} The five capability dimensions connect to 11 evaluation targets and 29 native metrics from 13 benchmark entries. Colored links trace the correspondence between dimensions, evaluation targets, metrics, and benchmark task settings. Metric tags identify rule/reference-based scoring (R), LLM-as-judge scoring (J), and detector-based model scoring (M). For each benchmark, $n$ denotes sampled units before expansion into evaluation cases.}
\label{fig:indieval_operationalization}
\end{figure}

Figure~\ref{fig:indieval_operationalization} summarizes how IndiEval connects the five FONTS dimensions to concrete evaluation settings. We organize these dimensions into 11 evaluation targets, operationalized through 29 native metrics across 13 benchmark entries. The mapping specifies which aspect of a capability each metric assesses within its benchmark's task setting, allowing a benchmark to contribute to multiple dimensions through distinct metrics.

To cover these evaluation targets, IndiEval brings together benchmarks spanning persona-conditioned decision-making, role-playing, user--assistant dialogue, narrative interaction, and social reasoning. Relative to Socio-Foundation's training, these benchmarks form two components: \textbf{(i) In-domain}, covering the task domains of the training datasets listed in Section~\ref{sec:implementation_details}; and \textbf{(ii) Out-of-distribution (OOD)}, consisting of AgentSense~\cite{mou2025agentsense} and UserLM-LiC~\cite{naous2026userlm,laban2026lic}. IndiEval retains the benchmarks' original evaluation protocols and organizes 29 task-specific metrics around these evaluation targets. We use DeepSeek-V4-Flash~\cite{deepseek2026v4flash} as both the auxiliary dialogue model and the judge model. Within a benchmark, different metrics may map to different capability dimensions, with detailed metric definitions and mappings provided in Appendix~\ref{app:indieval_metrics}. The overall Avg. includes both components and equally weights the five capability scores, each computed by averaging metrics within each dataset and then weighting datasets equally. Evaluation set sizes are provided in Appendix~\ref{app:evaluation_sampling}.

Figure~\ref{fig:indieval_distributions} summarizes the composition of IndiEval's 6,297 evaluation cases and their coverage of scoring types and capability dimensions. For the coverage summaries, each benchmark's case count is divided equally among its associated scoring types or dimensions. These descriptive proportions are separate from score aggregation: IndiEval Avg. gives equal weight to the five dimensions after averaging metrics within each dataset and datasets within each dimension.

\begin{figure}[!htbp]
\centering
\includegraphics[width=\linewidth]{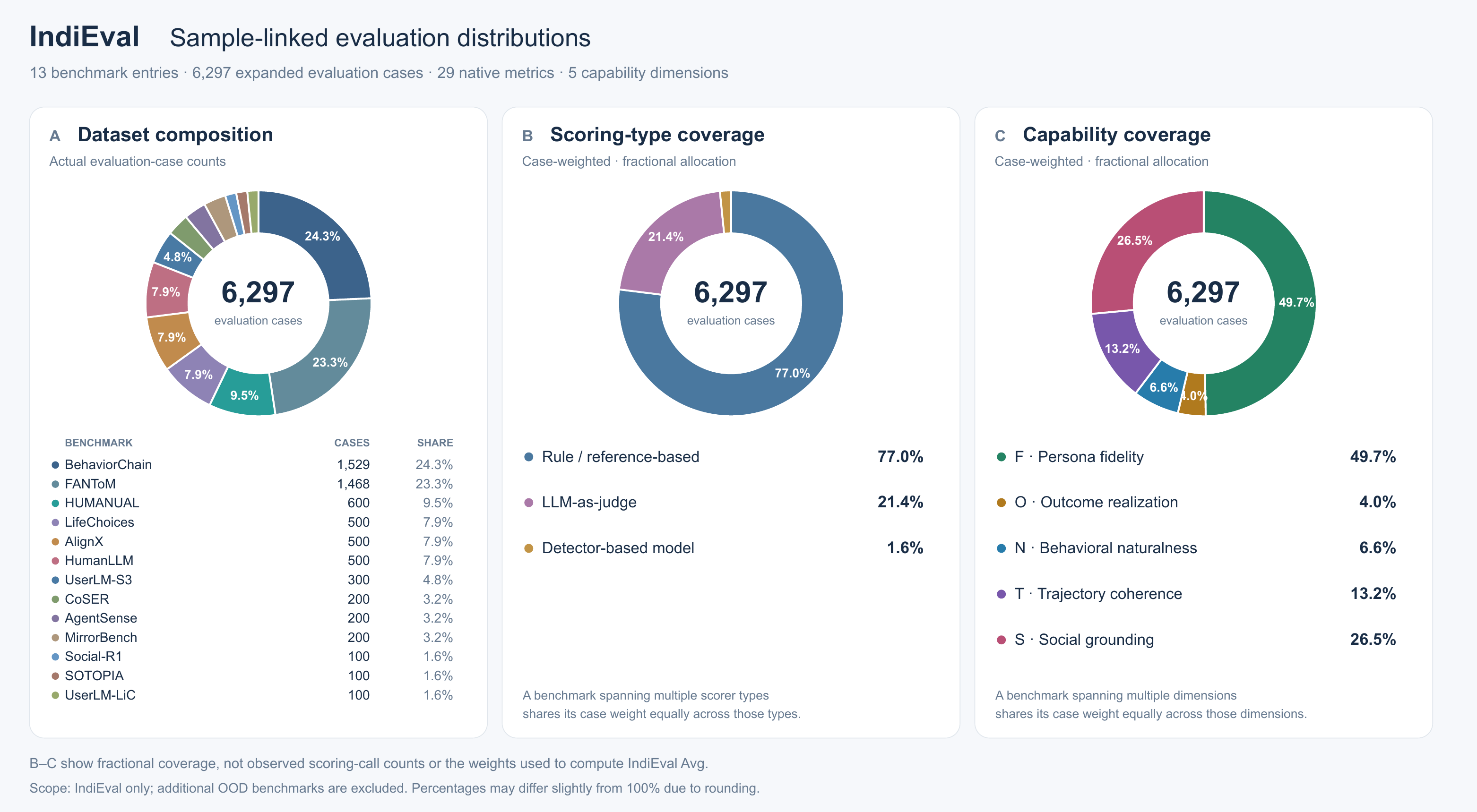}
\caption{\textbf{Evaluation-set composition and coverage of IndiEval.} (A) Distribution of 6,297 evaluation cases across 13 benchmark entries. (B) Scoring-type coverage and (C) capability coverage, weighted by evaluation-case counts. For benchmarks spanning multiple scoring types or capability dimensions, their case counts are allocated equally across the corresponding categories. These coverage proportions are independent of the weights used to compute IndiEval Avg. Additional OOD benchmarks outside IndiEval are excluded.}
\label{fig:indieval_distributions}
\end{figure}

\paragraph{Additional OOD evaluation beyond IndiEval.}
To further assess generalization beyond IndiEval, we separately evaluate Mistakes~\cite{ross2025learningmakemistakesmodeling}, HiToM~\cite{wu-etal-2023-hi}, and $\tau$-USI~\cite{zhou2026mind}. These three benchmarks supplement the OOD component within IndiEval and are excluded from Table~\ref{tab:main} scores and Avg. 
\begin{table}[t]
\centering
\caption{\textbf{Pipeline ablations on IndiEval.} We compare the full pipeline with SFT on task-expert rollouts, Task-expert OPD, LoRA averaging of task or capability experts, final-stage off-policy distillation from capability experts, SFT on pooled task data, and OPD with reverse KL. \textbf{Bold} and \underline{underline} indicate the best and second-best distinct scores in each column.}
\label{tab:ablation}
\begingroup
\definecolor{ablationheader}{RGB}{242,244,246}
\definecolor{ablationours}{RGB}{235,244,250}
\setlength{\tabcolsep}{1.0pt}
\renewcommand{\arraystretch}{1.16}
\scriptsize
\resizebox{\linewidth}{!}{%
\begin{tabular}{@{}lcccccccccccccccccccccccccccccc@{}}
\toprule
\rowcolor{ablationheader}
 & \multicolumn{9}{c}{\textbf{F}} & \multicolumn{7}{c}{\textbf{O}} & \multicolumn{7}{c}{\textbf{N}} & \multicolumn{3}{c}{\textbf{T}} & \multicolumn{3}{c}{\textbf{S}} & \textbf{Avg.} \\
\cmidrule(lr){2-10} \cmidrule(lr){11-17} \cmidrule(lr){18-24} \cmidrule(lr){25-27} \cmidrule(lr){28-30} \cmidrule(lr){31-31}
\rowcolor{ablationheader}
\textbf{Variant} & \metricheadtwo{LifeChoices}{Acc. $\uparrow$} & \metricheadtwo{AlignX}{Choice acc. $\uparrow$} & \metricheadtwo{HumanLLM}{Hit@5 $\uparrow$} & \metricheadtwo{BehaviorChain}{Node $\uparrow$} & \metricheadtwo{CoSER}{Char. fid. $\uparrow$} & \metricheadtwo{HUMANUAL}{Response $\uparrow$} & \metricheadtwo{HUMANUAL}{State $\uparrow$} & \metricheadtwo{UserLM-S3}{Intent adh. $\uparrow$} & \metricheadtwo{SOTOPIA}{Believ. $\uparrow$} & \metricheadtwo{SOTOPIA}{Goal $\uparrow$} & \metricheadtwo{SOTOPIA}{Relation. $\uparrow$} & \metricheadtwo{SOTOPIA}{Knowledge $\uparrow$} & \metricheadtwo{SOTOPIA}{Benefits $\uparrow$} & \metricheadtwo{AgentSense}{Goal maj. $\uparrow$} & \metricheadtwo{UserLM-LiC}{Intent cov. $\uparrow$} & \metricheadtwo{UserLM-LiC}{Task score $\uparrow$} & \metricheadtwo{UserLM-S3}{Term. F1 $\uparrow$} & \metricheadtwo{MirrorBench}{GTEval $\uparrow$} & \metricheadtwo{CoSER}{Anthrop. $\uparrow$} & \metricheadtwo{UserLM-S3}{Human lik. $\uparrow$} & \metricheadtwo{UserLM-S3}{Diversity $\uparrow$} & \metricheadtwo{UserLM-S3}{Inv. overlap $\uparrow$} & \metricheadtwo{UserLM-S3}{Role adh. $\uparrow$} & \metricheadtwo{CoSER}{Story cons. $\uparrow$} & \metricheadtwo{BehaviorChain}{CumScore $\uparrow$} & \metricheadtwo{CoSER}{Story qual. $\uparrow$} & \metricheadtwo{FANToM}{Item acc. $\uparrow$} & \metricheadtwo{Social-R1}{Acc. $\uparrow$} & \metricheadtwo{AgentSense}{Private acc. $\uparrow$} & \metrichead{Average $\uparrow$} \\
\midrule
Task-expert SFT & 72.8 & \underline{58.6} & \textbf{58.6} & 83.9 & \underline{26.3} & \underline{14.5} & \textbf{41.2} & \textbf{100.0} & \textbf{77.6} & 45.3 & \underline{55.6} & \textbf{28.3} & 48.7 & 91.2 & 70.0 & 31.6 & 45.3 & \underline{53.9} & 30.1 & 34.0 & 80.5 & \textbf{97.9} & \textbf{100.0} & 53.4 & 44.2 & 48.1 & 78.2 & 61.0 & 79.6 & \underline{59.5} \\
Task-expert OPD & 70.8 & 52.6 & 56.6 & 77.4 & 19.1 & 13.2 & 35.5 & \underline{99.0} & 72.6 & 44.4 & 54.8 & 23.0 & 50.0 & \underline{91.4} & \underline{74.6} & \underline{36.8} & 0.0 & 42.8 & 21.1 & 34.7 & 85.4 & 79.2 & \underline{91.0} & 51.7 & 43.2 & 38.8 & 60.2 & 48.0 & 82.1 & 54.2 \\
Task-expert averaging & 68.2 & 57.2 & 55.8 & 76.5 & 20.6 & 12.2 & 35.0 & 97.0 & 77.4 & 45.7 & 55.5 & \underline{27.1} & 50.7 & 89.5 & 72.0 & 32.3 & 12.1 & 40.2 & 25.4 & 26.4 & 86.1 & 74.5 & 71.0 & 49.6 & 44.8 & 42.2 & 63.9 & 50.0 & 83.5 & 54.5 \\
Capability-expert averaging & 70.2 & 55.4 & \underline{58.2} & 77.6 & 20.4 & 12.6 & 35.4 & \textbf{100.0} & 76.5 & \textbf{46.5} & \textbf{55.9} & 25.3 & 50.5 & 89.1 & \textbf{79.2} & \textbf{37.9} & \textbf{58.1} & 42.1 & 26.5 & 27.1 & 82.9 & 92.2 & \textbf{100.0} & 53.7 & 45.3 & 44.2 & 65.8 & 51.0 & 79.6 & 56.8 \\
Capability-expert off-policy distill. & 75.2 & 55.0 & 53.6 & \textbf{87.3} & 23.6 & 13.2 & 38.1 & \textbf{100.0} & 70.3 & 45.5 & 52.4 & 25.0 & 49.8 & 89.5 & 55.5 & 20.4 & \underline{54.8} & 50.6 & 29.1 & 40.4 & 79.8 & 96.1 & \textbf{100.0} & \underline{56.0} & \textbf{54.1} & 47.7 & 72.3 & \textbf{65.0} & \textbf{85.1} & 59.3 \\
Pooled-data SFT & \underline{77.4} & 57.8 & 40.6 & 77.9 & \textbf{42.3} & 6.3 & 19.5 & \textbf{100.0} & 32.6 & 25.7 & 34.2 & 6.0 & 48.4 & 4.3 & 31.2 & 23.5 & 12.9 & 28.5 & \textbf{33.8} & \textbf{73.2} & \textbf{93.0} & 51.6 & \textbf{100.0} & \textbf{64.4} & 34.8 & \textbf{55.2} & 61.7 & 57.0 & 29.6 & 43.0 \\
Reverse KL OPD & 73.8 & 57.8 & 54.8 & 83.1 & 17.8 & 13.0 & \underline{39.5} & 91.0 & 16.6 & 10.9 & 31.1 & 2.2 & 48.6 & 91.2 & 0.0 & 0.0 & 7.7 & 48.4 & 24.9 & 37.7 & 89.5 & 95.6 & 77.0 & 48.8 & 45.5 & 43.0 & \underline{79.4} & \underline{64.0} & 75.5 & 50.9 \\
Qwen3-8B & 72.2 & 57.0 & 56.4 & 34.9 & 21.0 & \textbf{14.6} & 38.7 & 93.0 & \underline{77.5} & \underline{46.4} & 54.2 & \underline{27.1} & \underline{51.6} & 90.8 & 70.7 & 29.3 & 12.9 & 40.1 & 25.5 & 25.4 & 87.4 & 75.9 & 70.0 & 54.0 & 6.9 & 44.3 & 75.2 & 50.0 & 80.5 & 50.7 \\
\midrule
\rowcolor{ablationours}
\textbf{Socio-Foundation} & \textbf{80.0} & \textbf{61.0} & \underline{58.2} & \underline{86.40} & 25.0 & 13.4 & \underline{39.5} & \textbf{100.0} & 74.4 & 45.2 & 54.2 & 26.5 & \textbf{52.9} & \textbf{93.1} & 72.8 & 34.7 & 52.5 & \textbf{54.1} & \underline{31.6} & \underline{47.9} & \underline{90.2} & \underline{97.1} & \textbf{100.0} & 54.7 & \underline{49.35} & \underline{48.6} & \textbf{79.6} & \underline{64.0} & \underline{83.7} & \textbf{61.7} \\
\bottomrule
\end{tabular}%
}
\endgroup

\end{table}

\paragraph{Models and comparisons.}
We compare Socio-Foundation with the proprietary \textit{GPT-5.5}~\cite{openai2026gpt55}; the open-weight \textit{DeepSeek-V4-Flash}~\cite{deepseek2026v4flash}, \textit{GLM-5.2}~\cite{zai2026glm52}, \textit{Kimi-K2.6}~\cite{moonshot2026kimik26}, \textit{Qwen3-Next-80B}~\cite{qwen2025next80b}, \textit{Qwen3-235B-A22B}~\cite{qwen2025instruct2507}, and \textit{Qwen3-8B}~\cite{qwen3technicalreport}; and the simulation-specialized \textit{OSim 8B}~\cite{zhou2026odyssim}, \textit{CoSER 8B}~\cite{wang2025coser}, \textit{HumanLM 8B}~\cite{wu2026humanlm}, \textit{UserLM 8B}~\cite{naous2026userlm}, and \textit{USP 8B}~\cite{wang2025usp}. %

\subsection{Main results}
\label{sec:main_results}

\paragraph{Overall performance.}

Table~\ref{tab:main} reports the performance of Socio-Foundation and the baseline models on IndiEval. Compared with the \textit{Qwen3-8B} base model, \textbf{Socio-Foundation improves the average score across the five dimensions by 11.0 points, with gains on 24 out of 29 metrics.} These results demonstrate that individual simulation capabilities can be systematically improved through our three-stage post-training pipeline. Socio-Foundation also remains among the leading specialized simulation models in terms of average performance. Despite its 8B parameter scale, it achieves overall performance comparable to that of several frontier general-purpose LLMs, such as \textit{GLM-5.2}, while exhibiting distinct strengths in specific behavioral capabilities.

\paragraph{Comparison with strong general-purpose LLMs.}
At the aggregate level, Socio-Foundation substantially outperforms the larger Qwen3 models, \textit{DeepSeek-V4-Flash}, and slightly outperforms \textit{Kimi-K2.6}, while remaining close to \textit{GLM-5.2} and \textit{GPT-5.5}. 
\textbf{In particular, general-purpose models are consistently weak on the N dimension}, where our Socio-Foundation performs better on 4 out of 7 metrics. These general-purpose models often drift back into the assistant role rather than role-playing as a natural individual, providing overly detailed information and sustaining the exchange rather than ending the conversation as naturally as human interlocutors do. This may reflect the strong assistant bias introduced during general-purpose post-training, which can make faithful user simulation more difficult.

\begin{figure}[!t]
\centering
\begin{minipage}[t]{0.32\linewidth}
\vspace{0pt}\centering
\includegraphics[width=\linewidth]{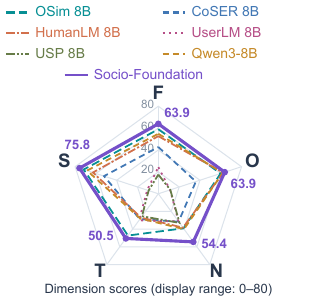}
\caption{\raggedright \textbf{Five-dimensional profiles of seven 8B models.}}
\label{fig:capability_radar}
\end{minipage}\hfill
\begin{minipage}[t]{0.32\linewidth}
\vspace{0pt}\centering
\includegraphics[width=\linewidth]{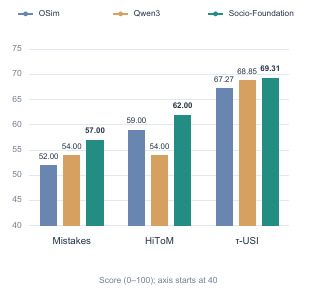}
\caption{\textbf{OOD evaluation. Mistakes, HiToM, and $\tau$-USI.}}
\label{fig:ood}
\end{minipage}\hfill
\begin{minipage}[t]{0.32\linewidth}
\vspace{0pt}\centering
\includegraphics[width=\linewidth]{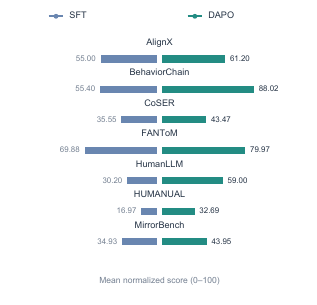}
\caption{\raggedright \textbf{SFT vs. DAPO task experts in several tasks.}}
\label{fig:task_expert_comparison}
\end{minipage}
\end{figure}

\paragraph{Breadth versus specialized simulators.}

Compared with similarly sized specialized models, \textbf{Socio-Foundation achieves the best or second-best result on 86\% of the metrics and the highest overall average score.} Existing specialized simulators often excel in the tasks for which they were designed, but generalize unevenly to other tasks. For example, \textit{UserLM} and \textit{USP} perform strongly on dialogue-oriented simulation tasks but transfer less effectively to decision-oriented evaluations, with reported scores falling to zero on some metrics. We observe that these narrowly specialized models can fail to follow instructions and carry out the requested task when evaluated outside their training task families. This limited transfer may reflect overfitting to dialogue-centric training data, which could constrain their ability to adapt to new task requirements. Socio-Foundation’s more consistent performance suggests that it consolidates individual-simulation capabilities across heterogeneous task formats rather than optimizing primarily for a particular interaction setting.

\subsection{Pipeline ablations}
\label{sec:pipeline_ablations}

Table~\ref{tab:ablation} evaluates seven alternatives to our hierarchical consolidation pipeline: (i) \textbf{Task-expert SFT}: task-expert rollouts are used to construct input-response pairs for training a single model; (ii) \textbf{Task-expert OPD}: directly using all task experts as teachers to do the MOPD aggregation. (iii)\textbf{Task-expert averaging}: the LoRA parameters of task experts are averaged; (iv) \textbf{Capability-expert averaging}: the LoRA parameters of capability experts are averaged; (v) \textbf{Capability-expert off-policy distillation}: final-stage on-policy distillation is replaced with off-policy trajectory distillation from capability experts; (vi) \textbf{Pooled-data SFT}: the \textit{Qwen3-8B} base model is directly fine-tuned on the pooled training data originally used to train the task experts; and (vii) \textbf{Reverse KL OPD}: the final on-policy integration stage replaces our sampled forward-KL surrogate, $D_{\mathrm{KL}}(q_d\|\pi_\theta)$, with reverse KL, $D_{\mathrm{KL}}(\pi_\theta\|q_d)$, while retaining the same capability-teacher integration structure.

\paragraph{Why use hierarchical integration and on-policy distillation?}
The full hierarchical pipeline achieves the highest Avg. in Table~\ref{tab:ablation} (61.7). Replacing final-stage on-policy distillation with \textit{Capability-expert off-policy distillation} reduces the Avg. to 59.3, while \textit{capability-expert averaging} further lowers it to 56.8, supporting the benefit of learned on-policy integration over the tested alternatives. The intermediate capability level also provides a consistent advantage: under the same parameter-averaging strategy, \textit{capability-expert averaging} outperforms \textit{task-expert averaging} (56.8 vs.\ 54.5), while directly integrating through \textit{Task-experts SFT} or \textit{Task-expert OPD} reaches 59.5 and 54.2 but remains below the full hierarchy, suggesting that direct integration can encounter conflicting optimization objectives. \textit{Pooled-data SFT} and \textit{Reverse KL OPD} obtain 43.0 and 50.9, respectively. The latter retains on-policy prefixes but reverses the KL direction, providing an objective-direction ablation for our forward-KL design. Together, these results support both stages of our design: consolidating task-level expertise into capability experts and integrating these experts through on-policy distillation yields the strongest unified model.

\paragraph{Why train task experts with DAPO?}
As shown in Figure~\ref{fig:task_expert_comparison}, DAPO task experts generally outperform their SFT counterparts across tasks based on dataset-level scores aggregated over normalized metrics, motivating the use of RL for task-expert training. We compare alternative RL methods in Appendix~\ref{app:whydapo}. We therefore adopt DAPO for task-expert training.%

\subsection{Mechanism and generalization analyses}

\paragraph{Additional OOD generalization beyond IndiEval.}
\label{sec:ood}
Beyond the OOD component within IndiEval, we evaluate Mistakes, HiToM, and $\tau$-USI as three additional benchmarks. Figure~\ref{fig:ood} compares \textit{OSim 8B}, \textit{Qwen3-8B}, and Socio-Foundation on the three benchmarks. Socio-Foundation obtains 57.00 on Mistakes, 62.00 on HiToM, and 69.31 on $\tau$-USI, outperforming both baselines on all three tasks and providing further evidence demonstrating the generalization and transability.

\paragraph{Five-dimensional capability profile.}
Figure~\ref{fig:capability_radar} presents five-dimensional scores aggregated from Table~\ref{tab:main}. The specialized models exhibit different strengths across dimensions. Socio-Foundation exceeds the five 8B specialized baselines and the \textit{Qwen3-8B base} model across all five dimensions, with its largest gains over the base model in trajectory coherence and behavioral naturalness, suggesting a stronger capacity for maintaining consistent behavior over extended interactions and producing more human-like responses.

\par\noindent
\begin{minipage}{\linewidth}
\begingroup
\setlength{\intextsep}{6pt}
\begin{wraptable}{l}{0.47\textwidth}
\centering\scriptsize
\setlength{\tabcolsep}{1.4pt}
\renewcommand{\arraystretch}{1.3}
\begin{tabular*}{\linewidth}{@{\extracolsep{\fill}}lcccccc@{}}
\toprule
Variant & \textbf{F} & \textbf{O} & \textbf{N} & \textbf{T} & \textbf{S} & \textbf{Avg.} \\
\midrule
\textbf{Socio-Foundation} & \textbf{63.93} & 63.85 & \textbf{54.41} & \textbf{50.50} & \textbf{75.77} & \textbf{61.69} \\
\midrule
w/o F & 60.72 & \textbf{66.99} & 53.33 & 49.38 & 72.81 & 60.65 \\
w/o O & 62.57 & 63.28 & 52.46 & 49.32 & 73.95 & 60.32 \\
w/o N & 61.66 & 64.12 & 46.66 & 49.19 & 74.59 & 59.24 \\
w/o T & 62.44 & 65.20 & 51.57 & 49.36 & 74.11 & 60.54 \\
w/o S & 62.94 & 66.03 & 51.87 & 48.93 & 68.85 & 59.72 \\
\bottomrule
\end{tabular*}
\caption{\textbf{Capability-expert ablations. w/o X denotes using other 4 experts to MOPD.} }
\label{tab:missing_capability}
\end{wraptable}
\paragraph{What is gained by integrating all five capability experts?}
As shown in Table~\ref{tab:missing_capability}, removing any capability expert lowers both the overall Avg. and the score on its corresponding capability dimension. Removing the behavioral naturalness expert (N) causes the largest decline in both overall performance and the corresponding capability score.
Removing F, N, T, or S raises O while lowering the overall Avg., suggesting a trade-off between outcome realization and the other capabilities in these configurations. These results show that the full five-expert configuration yields the strongest overall capability profile, supporting the complementary contributions of capability-level supervision to the unified model.
\par
\endgroup
\end{minipage}
\par

\section{Conclusion}

In this work, we introduce Socio-Foundation, a foundation model for simulating individual behaviors across diverse social contexts. We organize the fragmented task landscape into five foundational capabilities and develop a three-stage hierarchical expert-distillation pipeline to consolidate task-specific expertise. We also introduce IndiEval to assess these capabilities while preserving task-specific evaluation protocols. Socio-Foundation improves over its Qwen3-8B base across all five capabilities, achieves the strongest overall performance among same-scale specialized simulators, and approaches frontier general-purpose models. Ablations show that on-policy integration outperforms direct expert merging and the tested off-policy alternative, while all five capability experts contribute to the strongest overall result, supporting the FONTS framework as an effective basis for organizing and integrating complementary behavioral capabilities. Gains on three out-of-distribution benchmarks further support generalization beyond the main evaluation suite. These findings show the value of capability-structured training for individual simulation.

\bibliography{references}
\bibliographystyle{assets/plainnat}

\appendix
\clearpage
\begingroup
\hypersetup{hidelinks}
\section{Dataset Survey, Capability Annotations, and Clustering}
\label{app:dataset_taxonomy}

This appendix documents the dataset-level evidence underlying the FONTS organization. The inventory contains 44 surveyed dataset/task entries and 10 supplementary benchmark or protocol entries used in the paper. Subsets and local derivatives are recorded separately when their simulated roles or task requirements differ; these 54 entries therefore do not represent 54 independent dataset families. This survey inventory is distinct from the selected training corpus.

\subsection{Annotation unit and organization}
For each dataset, we record the task formulation, the simulated or inferred subject, the relevant scenario, and the behavioral requirement suggested by its evaluation protocol. The annotation unit is a \emph{dataset--task--role configuration}: the same source can support different requirements under different conversions. We represent these requirements with 19 labels (A01--A19) and organize them into eight semantic clusters (C1--C8). The theory-guided consolidation described in the main text maps C1--C3 to F, C4 to O, C5 to N, C6 to T, and C7--C8 to S. Table~\ref{tab:fonts_cluster_dictionary} specifies the complete mapping.

The assignments describe capability demands rather than mutually exclusive task categories or independent scores for every label. A requirement can be inferred from a task's scenario without having a dedicated native metric. Local transformations and role-specific interpretations remain qualified in the inventory. In particular, a multi-turn format alone does not establish T; maintaining a supplied user's intent supports F, whereas achieving that user's desired result supports O. Interpreting another person's intent supports S.

\begin{table}[!ht]
\centering
\caption{The eight capability clusters and their 19 requirement labels. F/O/N/T/S denote Persona Fidelity, Outcome Realization, Behavioral Naturalness, Trajectory Coherence, and Social Grounding.}
\label{tab:fonts_cluster_dictionary}
\small
\setlength{\tabcolsep}{4pt}
\renewcommand{\arraystretch}{1.18}
\begin{tabular}{@{}>{\raggedright\arraybackslash}p{0.075\linewidth}>{\raggedright\arraybackslash}p{0.285\linewidth}>{\raggedright\arraybackslash}p{0.60\linewidth}@{}}
\toprule
\textbf{Cluster} & \textbf{Theme / dimension} & \textbf{Capability requirements} \\
\midrule
C1 & Preference / Value Fidelity (\textbf{F}) & \textbf{A01} Preference alignment;\newline \textbf{A02} Viewpoint and value fidelity \\
C2 & Persona / Expression Fidelity (\textbf{F}) & \textbf{A03} Persona maintenance;\newline \textbf{A04} Individual expression-style fidelity \\
C3 & Behavioral / Decision Fidelity (\textbf{F}) & \textbf{A05} Individual response and choice fidelity;\newline \textbf{A06} User-intent fidelity;\newline \textbf{A10} Individual cognitive and decision-pattern fidelity \\
C4 & Goal / Outcome Realization (\textbf{O}) & \textbf{A07} Individual goal realization \\
C5 & Human-like Expression / Behavior (\textbf{N}) & \textbf{A08} Natural dialogue behavior;\newline \textbf{A09} Human-like behavior \\
C6 & Trajectory / Narrative Coherence (\textbf{T}) & \textbf{A11} Behavior-sequence and cross-turn consistency;\newline \textbf{A12} Narrative progression and response consistency \\
C7 & Mental-State / Perspective Grounding (\textbf{S}) & \textbf{A13} Others' beliefs and knowledge;\newline \textbf{A14} Perspective and information boundaries;\newline \textbf{A19} Intention and motive understanding \\
C8 & Social / Emotional Context Grounding (\textbf{S}) & \textbf{A15} Emotion and situation understanding;\newline \textbf{A16} Social relationships and roles;\newline \textbf{A17} Social norms and behavioral consequences;\newline \textbf{A18} Context-appropriate emotional responses \\
\bottomrule
\end{tabular}
\end{table}
\clearpage

\subsection{Overview of dataset coverage}
Figure~\ref{fig:fonts_dataset_clusters} visualizes the complete inventory. Repeated dataset labels indicate multiple annotated requirements. BehaviorChain appears in C3/F for persona-conditioned node behavior and C6/T for progression across a behavior chain. UserLM-S3 and UserLM-LiC are separated because the former evaluates intent fidelity and natural user behavior, while the latter evaluates intent coverage and task outcomes. The supplementary Mistakes entry concerns reproducing a supplied student's misconception and is assigned to C3/F. HiToM, $\tau$-USI, and Mistakes remain the separate OOD benchmarks used in the paper; their inclusion in this survey does not imply training use.

\begin{figure}[!ht]
\centering
\includegraphics[width=\linewidth]{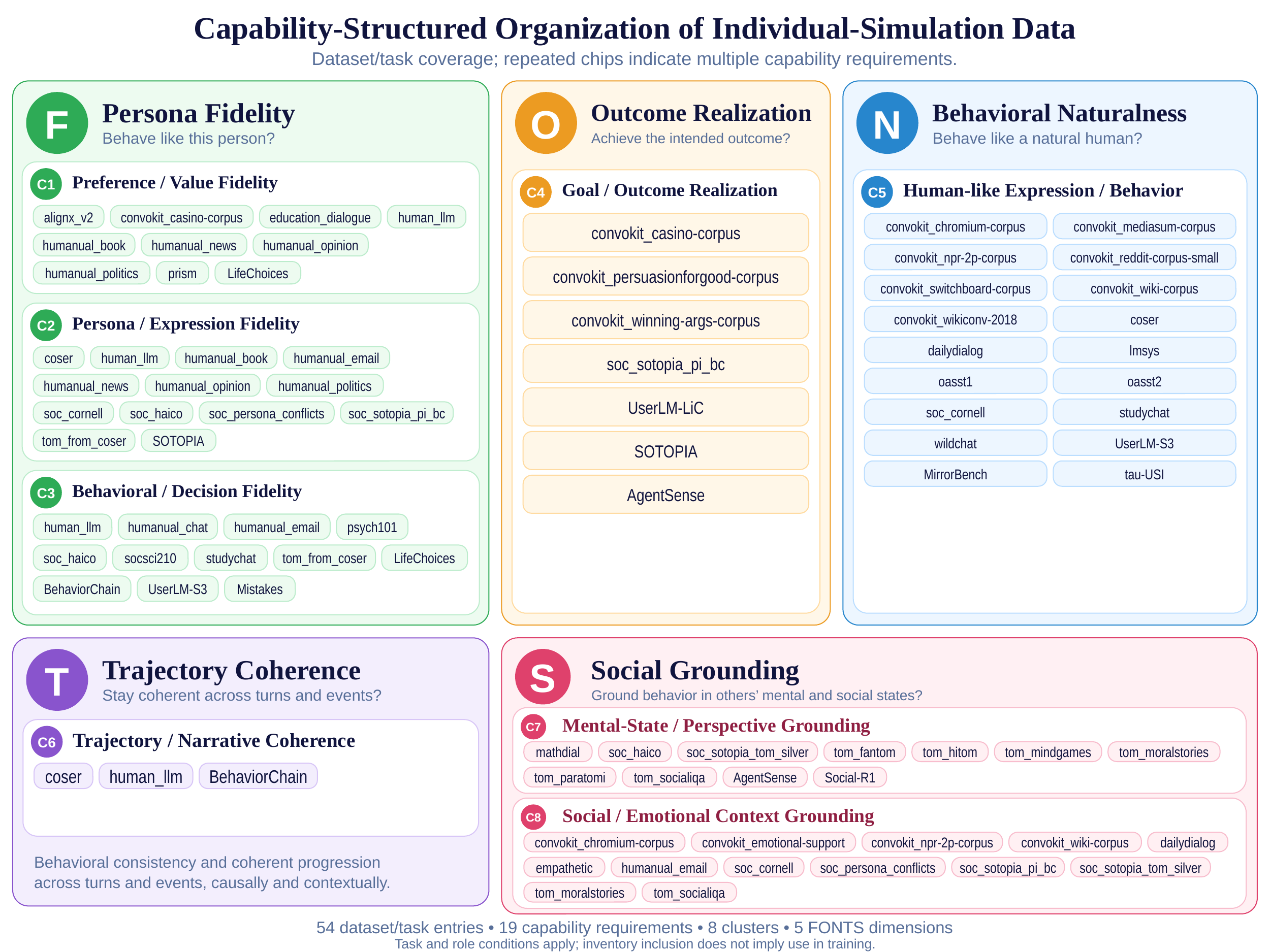}
\caption{\textbf{Capability-structured organization of the surveyed data and supplementary benchmarks.} The five colored regions contain eight clusters. Each chip identifies a dataset/task entry, and an entry may appear in multiple clusters. The diagram summarizes the annotations in Table~\ref{tab:fonts_dataset_inventory}; the stated task and role interpretations apply to the displayed memberships. Table~\ref{tab:fonts_metric_concordance} separately records the dimensions measured by the selected main-table metrics.}
\label{fig:fonts_dataset_clusters}
\end{figure}

\clearpage
\subsection{Detailed inventory and annotation results}
Table~\ref{tab:fonts_dataset_inventory} presents the dataset/task names, task interpretations, capability requirements, clusters, and FONTS dimensions in separate columns. Dataset names link to the public task or parent source. A dagger ($\dagger$) marks a local identifier or derivative whose exact conversion is not established by the public parent source. Assignments refer to the stated task and role interpretations; additional tags are not inferred solely from a dataset name or a multi-turn format.

\begingroup
\fontsize{8}{9.6}\selectfont
\setlength{\tabcolsep}{2.4pt}
\renewcommand{\arraystretch}{1.13}
\begin{longtable}{@{}>{\raggedright\arraybackslash}p{0.305\linewidth}>{\raggedright\arraybackslash}p{0.295\linewidth}>{\raggedright\arraybackslash}p{0.135\linewidth}>{\centering\arraybackslash}p{0.105\linewidth}>{\centering\arraybackslash}p{0.11\linewidth}@{}}
\caption{Dataset/task inventory and capability annotations. A-labels are defined in Table~\ref{tab:fonts_cluster_dictionary}; C-labels identify clusters. The final two columns report cluster membership and FONTS dimensions separately. Supplementary entries complete coverage of the paper's benchmarks and protocols.}\label{tab:fonts_dataset_inventory}\\
\toprule
\textbf{Dataset / task} & \textbf{Task interpretation} & \textbf{Requirements} & \textbf{Clusters} & \textbf{Dimensions} \\
\midrule
\endfirsthead
\multicolumn{5}{l}{\textit{Table \thetable\ continued}}\\
\toprule
\textbf{Dataset / task} & \textbf{Task interpretation} & \textbf{Requirements} & \textbf{Clusters} & \textbf{Dimensions} \\
\midrule
\endhead
\midrule
\multicolumn{5}{r}{\textit{Continued on next page}}\\
\endfoot
\bottomrule
\endlastfoot
\multicolumn{5}{@{}l}{\textbf{Surveyed dataset/task entries}}\\[3pt]
\href{https://github.com/JinaLeejnl/AlignX}{\texttt{alignx\_\allowbreak v2}}$^{\dagger}$ & Preference-conditioned response selection. & A01 & C1 & \textbf{F} \\[3pt]
\href{https://convokit.cornell.edu/documentation/casino-corpus.html}{\texttt{convokit\_\allowbreak casino-\allowbreak corpus}} & Negotiation with private item preferences and individual objectives. & A01, A07 & C1, C4 & \textbf{F, O} \\[3pt]
\href{https://convokit.cornell.edu/documentation/chromium.html}{\texttt{convokit\_\allowbreak chromium-\allowbreak corpus}} & Participants in software-review discussions. & A08, A16 & C5, C8 & \textbf{N, S} \\[3pt]
\href{https://convokit.cornell.edu/documentation/support.html}{\texttt{convokit\_\allowbreak emotional-\allowbreak support}} & Emotional-support conversations. & A15, A18 & C8 & \textbf{S} \\[3pt]
\href{https://convokit.cornell.edu/documentation/mediasum.html}{\texttt{convokit\_\allowbreak mediasum-\allowbreak corpus}} & Dialogue simulation from interview transcripts. & A08 & C5 & \textbf{N} \\[3pt]
\href{https://convokit.cornell.edu/documentation/npr-2p.html}{\texttt{convokit\_\allowbreak npr-\allowbreak 2p-\allowbreak corpus}} & Two-person radio interviews. & A08, A16 & C5, C8 & \textbf{N, S} \\[3pt]
\href{https://convokit.cornell.edu/documentation/persuasionforgood.html}{\texttt{convokit\_\allowbreak persuasion\allowbreak forgood-\allowbreak corpus}} & Donation persuasion with an explicit participant goal. & A07 & C4 & \textbf{O} \\[3pt]
\href{https://convokit.cornell.edu/documentation/reddit-small.html}{\texttt{convokit\_\allowbreak reddit-\allowbreak corpus-\allowbreak small}} & Participants in online discussion threads. & A08 & C5 & \textbf{N} \\[3pt]
\href{https://convokit.cornell.edu/documentation/switchboard.html}{\texttt{convokit\_\allowbreak switchboard-\allowbreak corpus}} & Telephone conversations. & A08 & C5 & \textbf{N} \\[3pt]
\href{https://convokit.cornell.edu/documentation/wiki.html}{\texttt{convokit\_\allowbreak wiki-\allowbreak corpus}} & Editors participating in Wikipedia discussions. & A08, A16 & C5, C8 & \textbf{N, S} \\[3pt]
\href{https://convokit.cornell.edu/documentation/wikiconv.html}{\texttt{convokit\_\allowbreak wikiconv-\allowbreak 2018}} & Wikipedia discussion histories. & A08 & C5 & \textbf{N} \\[3pt]
\href{https://convokit.cornell.edu/documentation/winning.html}{\texttt{convokit\_\allowbreak winning-\allowbreak args-\allowbreak corpus}} & Argumentative dialogue aimed at changing a view. & A07 & C4 & \textbf{O} \\[3pt]
\href{https://arxiv.org/html/2502.09082v1}{\texttt{coser}} & Established literary characters interacting in scenes. & A03, A09, A12 & C2, C5, C6 & \textbf{F, N, T} \\[3pt]
\href{https://aclanthology.org/I17-1099/}{\texttt{dailydialog}} & Everyday dialogue with emotional context. & A08, A18 & C5, C8 & \textbf{N, S} \\[3pt]
\href{https://github.com/google-research-datasets/Education-Dialogue-Dataset}{\texttt{education\_\allowbreak dialogue}} & Educational dialogue with stated preferences. & A01 & C1 & \textbf{F} \\[3pt]
\href{https://aclanthology.org/P19-1534/}{\texttt{empathetic}} & Responses grounded in an interlocutor's emotional experience. & A15, A18 & C8 & \textbf{S} \\[3pt]
\href{https://arxiv.org/html/2601.15793v1}{\texttt{human\_\allowbreak llm}} & Personalized selection, expression, and scenario/story subtasks. & A01, A04, A05, A12 & C1, C2, C3, C6 & \textbf{F, T} \\[3pt]
\href{https://arxiv.org/html/2603.03303v1}{\texttt{humanual\_\allowbreak book}} & Book-related responses conditioned on user information. & A01, A04 & C1, C2 & \textbf{F} \\[3pt]
\href{https://arxiv.org/html/2603.03303v1}{\texttt{humanual\_\allowbreak chat}} & Chat responses conditioned on a user's intent and state. & A05, A06 & C3 & \textbf{F} \\[3pt]
\href{https://arxiv.org/html/2603.03303v1}{\texttt{humanual\_\allowbreak email}} & Email responses conditioned on the user and relationship context. & A04, A05, A16 & C2, C3, C8 & \textbf{F, S} \\[3pt]
\href{https://arxiv.org/html/2603.03303v1}{\texttt{humanual\_\allowbreak news}} & News-related personalized responses. & A02, A04 & C1, C2 & \textbf{F} \\[3pt]
\href{https://arxiv.org/html/2603.03303v1}{\texttt{humanual\_\allowbreak opinion}} & Personalized responses about opinions. & A02, A04 & C1, C2 & \textbf{F} \\[3pt]
\href{https://arxiv.org/html/2603.03303v1}{\texttt{humanual\_\allowbreak politics}} & Personalized political responses. & A02, A04 & C1, C2 & \textbf{F} \\[3pt]
\href{https://huggingface.co/datasets/lmsys/lmsys-chat-1m}{\texttt{lmsys}} & User-side conversational simulation. & A08 & C5 & \textbf{N} \\[3pt]
\href{https://arxiv.org/abs/2305.14536}{\texttt{mathdial}} & Teacher responses grounded in a learner's knowledge state. & A13 & C7 & \textbf{S} \\[3pt]
\href{https://huggingface.co/datasets/OpenAssistant/oasst1}{\texttt{oasst1}} & Dialogue-role simulation from conversation trees. & A08 & C5 & \textbf{N} \\[3pt]
\href{https://huggingface.co/datasets/OpenAssistant/oasst2}{\texttt{oasst2}} & Dialogue-role simulation from conversation trees. & A08 & C5 & \textbf{N} \\[3pt]
\href{https://github.com/HannahKirk/prism-alignment}{\texttt{prism}} & Participant-conditioned preference judgments. & A01 & C1 & \textbf{F} \\[3pt]
\href{https://huggingface.co/datasets/marcelbinz/Psych-101}{\texttt{psych101}} & Participant-conditioned cognitive and decision tasks. & A10 & C3 & \textbf{F} \\[3pt]
\href{https://convokit.cornell.edu/documentation/movie.html}{\texttt{soc\_\allowbreak cornell}}$^{\dagger}$ & Character dialogue adapted from movie conversations. & A03, A09, A16 & C2, C5, C8 & \textbf{F, N, S} \\[3pt]
\href{https://arxiv.org/html/2409.16427v1}{\texttt{soc\_\allowbreak haico}}$^{\dagger}$ & Persona/intent-conditioned interaction with information boundaries. & A03, A06, A14 & C2, C3, C7 & \textbf{F, S} \\[3pt]
\href{https://aclanthology.org/2025.emnlp-main.586/}{\texttt{soc\_\allowbreak persona\_\allowbreak conflicts}}$^{\dagger}$ & Persona-conditioned interaction in social conflict. & A03, A15, A16 & C2, C8 & \textbf{F, S} \\[3pt]
\href{https://arxiv.org/html/2310.11667v1}{\texttt{soc\_\allowbreak sotopia\_\allowbreak pi\_\allowbreak bc}}$^{\dagger}$ & SOTOPIA-derived role-play with goals and social relationships. & A03, A07, A16 & C2, C4, C8 & \textbf{F, O, S} \\[3pt]
\href{https://arxiv.org/html/2310.11667v1}{\texttt{soc\_\allowbreak sotopia\_\allowbreak tom\_\allowbreak silver}}$^{\dagger}$ & SOTOPIA-derived social and mental-state reasoning. & A13, A14, A16 & C7, C8 & \textbf{S} \\[3pt]
\href{https://arxiv.org/html/2509.05830v1}{\texttt{socsci210}} & Participant-conditioned social-science judgments. & A10 & C3 & \textbf{F} \\[3pt]
\href{https://arxiv.org/html/2503.07928v1}{\texttt{studychat}} & Student-side dialogue with an identifiable learning intent. & A08, A06 & C3, C5 & \textbf{F, N} \\[3pt]
\href{https://github.com/skywalker023/fantom}{\texttt{tom\_\allowbreak fantom}} & Reasoning about beliefs and knowledge in conversations. & A13, A14 & C7 & \textbf{S} \\[3pt]
\href{https://arxiv.org/html/2502.09082v1}{\texttt{tom\_\allowbreak from\_\allowbreak coser}}$^{\dagger}$ & CoSER-derived persona-conditioned responses. & A03, A05 & C2, C3 & \textbf{F} \\[3pt]
\href{https://github.com/ying-hui-he/Hi-ToM_dataset}{\texttt{tom\_\allowbreak hitom}}\newline\textit{Additional OOD} & Reasoning about nested beliefs and information access. & A13, A14 & C7 & \textbf{S} \\[3pt]
\href{https://huggingface.co/datasets/sileod/mindgames}{\texttt{tom\_\allowbreak mindgames}} & Reasoning about beliefs under perspective differences. & A13, A14 & C7 & \textbf{S} \\[3pt]
\href{https://huggingface.co/datasets/demelin/moral_stories}{\texttt{tom\_\allowbreak moralstories}} & Reasoning about motives, norms, and social consequences. & A17, A19 & C7, C8 & \textbf{S} \\[3pt]
\href{https://github.com/msclar/symbolictom}{\texttt{tom\_\allowbreak paratomi}} & Reasoning about perspectives in ToM stories. & A13, A14 & C7 & \textbf{S} \\[3pt]
\href{https://huggingface.co/datasets/allenai/social_i_qa}{\texttt{tom\_\allowbreak socialiqa}} & Reasoning about social motives and emotional situations. & A19, A15 & C7, C8 & \textbf{S} \\[3pt]
\href{https://huggingface.co/datasets/allenai/WildChat}{\texttt{wildchat}} & User-side simulation from in-the-wild dialogue. & A08 & C5 & \textbf{N} \\[3pt]
\midrule\multicolumn{5}{@{}l}{\textbf{Supplementary benchmark/protocol entries}}\\[3pt]
\href{https://aclanthology.org/2025.findings-emnlp.813/}{\texttt{LifeChoices}} & Persona-conditioned decisions in life situations. & A01, A05 & C1, C3 & \textbf{F} \\[3pt]
\href{https://aclanthology.org/2025.findings-acl.813/}{\texttt{BehaviorChain}} & Predict successive behaviors from persona history and an evolving scenario. & A05, A11 & C3, C6 & \textbf{F, T} \\[3pt]
\href{https://arxiv.org/abs/2510.06552}{\texttt{UserLM-\allowbreak S3}} & Simulate user dialogue given intent and conversational history. & A06, A08, A09 & C3, C5 & \textbf{F, N} \\[3pt]
\href{https://arxiv.org/abs/2510.06552}{\texttt{UserLM-\allowbreak LiC}}\newline\textit{Main-table OOD} & User simulation with progressively communicated task requirements. & A07 & C4 & \textbf{O} \\[3pt]
\href{https://arxiv.org/abs/2310.11667}{\texttt{SOTOPIA}} & Persona-conditioned social interaction with individual objectives. & A03, A07 & C2, C4 & \textbf{F, O} \\[3pt]
\href{https://aclanthology.org/2025.naacl-long.257/}{\texttt{AgentSense}}\newline\textit{Main-table OOD} & Multi-agent social scenarios with goals and private information. & A07, A13, A14 & C4, C7 & \textbf{O, S} \\[3pt]
\href{https://openreview.net/forum?id=WWYowYlG70}{\texttt{MirrorBench}} & Conversational user proxies evaluated for human-likeness. & A08, A09 & C5 & \textbf{N} \\[3pt]
\href{https://arxiv.org/abs/2603.09249}{\texttt{Social-\allowbreak R1}} & Reason about social mental states and perspectives. & A13, A14, A19 & C7 & \textbf{S} \\[3pt]
\href{https://arxiv.org/abs/2603.11245}{\texttt{tau-\allowbreak USI}}\newline\textit{Additional OOD} & User simulation in tool-mediated service tasks. & A09 & C5 & \textbf{N} \\[3pt]
\href{https://arxiv.org/abs/2510.11502}{\texttt{Mistakes}}\newline\textit{Additional OOD} & Predict the wrong answer produced under a supplied student misconception. & A05, A10 & C3 & \textbf{F} \\[3pt]
\end{longtable}
\endgroup

\subsection{Concordance with the evaluation dimensions}
\label{app:taxonomy_concordance}
Dataset-level requirements and selected evaluation metrics have different scopes. For example, the HumanLLM survey entry includes a story/scenario interpretation under T, while its main-table item-selection metric measures F. Likewise, the HUMANUAL email interpretation includes social context, but the selected HUMANUAL response/state alignment scores are assigned to F. These additional survey memberships do not introduce extra evaluation scores or change the five-dimension average. AgentSense and UserLM-LiC are OOD tasks within the main table, while HiToM, $\tau$-USI, and Mistakes form the separate additional OOD analysis.

Table~\ref{tab:fonts_metric_concordance} exhaustively aligns the 29 main-table metrics with the relevant clusters. It preserves the paper's metric assignments: SOTOPIA believability is F; UserLM-S3 user-role adherence and termination F1 are N; and the two UserLM-LiC metrics are O. The memberships shown for SOTOPIA and the added benchmark/protocol entries follow these selected task interpretations. Local SOTOPIA derivatives retain their own qualified survey annotations.

\begingroup
\small
\setlength{\tabcolsep}{4pt}
\renewcommand{\arraystretch}{1.17}
\begin{longtable}{@{}>{\raggedright\arraybackslash}p{0.195\linewidth}>{\raggedright\arraybackslash}p{0.175\linewidth}>{\raggedright\arraybackslash}p{0.59\linewidth}@{}}
\caption{Concordance between cluster memberships and the selected evaluation metrics. Each of the 29 main-table metrics appears once; this table introduces no new metric or score.}\label{tab:fonts_metric_concordance}\\
\toprule
\textbf{Benchmark / entry} & \textbf{Cluster / dimension} & \textbf{Selected metrics and interpretation}\\
\midrule
\endfirsthead
\multicolumn{3}{l}{\textit{Table \thetable\ continued}}\\
\toprule
\textbf{Benchmark / entry} & \textbf{Cluster / dimension} & \textbf{Selected metrics and interpretation}\\
\midrule
\endhead
\bottomrule
\endfoot
LifeChoices & C1, C3 / F & Choice accuracy: persona-conditioned preference and choice fidelity.\\
AlignX & C1 / F & Direct-choice accuracy: personalized preference fidelity.\\
HumanLLM & C1 / F & Hit@5: personalized item selection; the story/scenario membership in the survey is outside this metric.\\
BehaviorChain & C3 / F; C6 / T & Node accuracy (F); chain-level CumScore (T).\\
CoSER & C2 / F; C5 / N; C6 / T & Character fidelity (F); anthropomorphism (N); storyline consistency and storyline quality (T).\\
HUMANUAL & C1--C3 / F & Response alignment and state alignment. The six domain entries supply the relevant individual-reference requirements.\\
UserLM-S3 & C3 / F; C5 / N & Intent adherence (F); termination F1, human likelihood, first-turn diversity, inverse intent overlap, and user-role adherence (N).\\
SOTOPIA & C2 / F; C4 / O & Believability (F); goal achievement, relationship, knowledge, and financial/material benefits (O).\\
AgentSense & C4 / O; C7 / S & Goal completion (O); private-information accuracy (S).\\
UserLM-LiC & C4 / O & Intent coverage and assistant task score: realizing the intended downstream result.\\
MirrorBench & C5 / N & GTEval: conversational human-likeness.\\
FANToM & C7 / S & Item accuracy: reasoning about others' mental states and information access.\\
Social-R1 & C7 / S & Selected accuracy: social mental-state reasoning.\\
\midrule
\multicolumn{3}{@{}l}{\textbf{Separate OOD analysis (excluded from the main-table average)}}\\
HiToM & C7 / S & Higher-order belief and perspective reasoning.\\
$\tau$-USI & C5 / N & Alignment with human behavior/outcomes in service interactions.\\
Mistakes & C3 / F & Misconception-conditioned student-answer simulation.\\
\end{longtable}
\endgroup

The additions are grounded in the corresponding task definitions: LifeChoices~\citep{xu2025lifechoice}, BehaviorChain~\citep{li2025behaviorchain}, UserLM and LiC~\citep{naous2026userlm,laban2026lic}, SOTOPIA~\citep{zhou2024sotopia}, AgentSense~\citep{mou2025agentsense}, MirrorBench~\citep{hathidara2026mirrorbench}, Social-R1~\citep{wu2026socialr1}, $\tau$-USI~\citep{zhou2026mind}, and Mistakes~\citep{ross2025learningmakemistakesmodeling}. The FONTS assignments are our capability organization of these tasks, rather than category labels attributed to the original dataset authors.
\endgroup

\section{Data Cleaning}
\label{app:data_cleaning}

\subsection{Cleaning Method}
\label{app:cleaning_method}
We apply rule-based cleaning to the collected datasets, distinguishing between repairs to formatting or prompt--target consistency and filtering of unsuitable instances. The completed rules address construction artifacts, ambiguous supervision, role confusion, duplication, and excessive length. Dataset-level retained, removed, and transformed counts are reported in Section~\ref{app:cleaning_results}.

\subsection{Cleaning Procedure}
\label{app:cleaning_procedure}
\paragraph{Repairs.}
We apply three types of repair:
\begin{enumerate}
    \item \textbf{Spurious symbols and markers.} Remove extraneous special symbols and noisy markers introduced during data construction.
    \item \textbf{Template inconsistencies.} Resolve inconsistencies introduced by shared templates used to construct the data.
    \item \textbf{Prompt--target mismatches.} Align targets with the requested output format, including removing chain-of-thought text when the prompt requests only an answer and reconciling numeric versus alphabetic answer labels (e.g., a prompt requesting labels 1--3 paired with a target using A--C).
\end{enumerate}

\paragraph{Filtering.}
We remove instances that meet any of the following seven criteria:
\begin{enumerate}
    \item \textbf{Role confusion.} The target contains outputs that mix multiple speaker roles.
    \item \textbf{Profile-construction noise.} An AlignX user profile contains artifacts such as \texttt{Response 1} or \texttt{Response 2} introduced during profile construction.
    \item \textbf{Empty targets.} The target is empty.
    \item \textbf{Unexpanded questionnaire templates.} A SocSci210 instance contains an unexpanded template that makes the meaning of its target uninterpretable.
    \item \textbf{Conflicting or ambiguous answers.} The target contains contradictory answers or does not specify a unique answer.
    \item \textbf{Duplicate instances.} The instance duplicates another entry in the data.
    \item \textbf{Excessive length.} The instance exceeds 32,768 tokens.
\end{enumerate}

\subsection{Cleaning Results}
\label{app:cleaning_results}
Table~\ref{tab:data_cleaning_results} reports 45 sources: 19,678,715 input rows, 19,018,645 retained rows, and 660,070 removed rows. The total transformation count is 52,615. The \textit{Transformed} column reports instances modified by the repair rules, whereas \textit{Removed} reports instances excluded by the filtering rules described above.

\begingroup
\hypersetup{hidelinks}
\fontsize{7.5}{9}\selectfont
\setlength{\tabcolsep}{1.5pt}
\renewcommand{\arraystretch}{1.00}
\begin{longtable}{@{}>{\raggedright\arraybackslash}p{0.365\linewidth}>{\raggedleft\arraybackslash}p{0.14\linewidth}>{\raggedleft\arraybackslash}p{0.14\linewidth}>{\raggedleft\arraybackslash}p{0.105\linewidth}>{\raggedleft\arraybackslash}p{0.125\linewidth}>{\raggedleft\arraybackslash}p{0.085\linewidth}@{}}
\caption{Dataset cleaning statistics. Transformed counts repaired instances; Removed counts filtered-out instances. Share is the percentage of all retained rows. Transformation counts are reported separately and are not subtracted again from retained counts.}\label{tab:data_cleaning_results}\\
\toprule
\textbf{Dataset / source} & \textbf{Before} & \textbf{After} & \textbf{Removed} & \textbf{Transformed} & \textbf{Share (\%)} \\
\midrule
\endfirsthead
\multicolumn{6}{l}{\textit{Table \thetable\ continued}}\\
\toprule
\textbf{Dataset / source} & \textbf{Before} & \textbf{After} & \textbf{Removed} & \textbf{Transformed} & \textbf{Share (\%)} \\
\midrule
\endhead
\midrule
\multicolumn{6}{r}{\textit{Continued on next page}}\\
\endfoot
\bottomrule
\endlastfoot
\texttt{alignx\_\allowbreak v2} & 14,649,170 & 14,086,414 & 562,756 & 1,097 & 74.0663 \\
\texttt{socsci210} & 2,618,745 & 2,547,535 & 71,210 & 30,311 & 13.3949 \\
\texttt{human\_\allowbreak llm} & 1,316,829 & 1,316,811 & 18 & 535 & 6.9238 \\
\texttt{wildchat} & 165,346 & 157,220 & 8,126 & 3,726 & 0.8267 \\
\texttt{coser} & 114,831 & 114,831 & 0 & 8 & 0.6038 \\
\texttt{lmsys} & 79,929 & 79,908 & 21 & 167 & 0.4202 \\
\texttt{tom\_\allowbreak from\_\allowbreak coser} & 76,747 & 76,747 & 0 & 0 & 0.4035 \\
\texttt{convokit\_\allowbreak chromium-\allowbreak corpus} & 70,365 & 70,365 & 0 & 2 & 0.3700 \\
\texttt{psych101} & 57,653 & 56,805 & 848 & 0 & 0.2987 \\
\texttt{humanual\_\allowbreak news} & 49,148 & 45,894 & 3,254 & 0 & 0.2413 \\
\texttt{empathetic} & 45,871 & 45,871 & 0 & 0 & 0.2412 \\
\texttt{humanual\_\allowbreak politics} & 45,395 & 45,393 & 2 & 0 & 0.2387 \\
\texttt{humanual\_\allowbreak opinion} & 38,613 & 38,613 & 0 & 2,445 & 0.2030 \\
\texttt{convokit\_\allowbreak npr-\allowbreak 2p-\allowbreak corpus} & 37,532 & 37,513 & 19 & 32 & 0.1972 \\
\texttt{tom\_\allowbreak socialiqa} & 33,067 & 32,882 & 185 & 0 & 0.1729 \\
\texttt{humanual\_\allowbreak book} & 31,931 & 31,930 & 1 & 7 & 0.1679 \\
\texttt{education\_\allowbreak dialogue} & 28,026 & 28,026 & 0 & 0 & 0.1474 \\
\texttt{UserLM-\allowbreak S3} & 28,918 & 26,134 & 2,784 & 0 & 0.1374 \\
\texttt{dailydialog} & 24,448 & 24,448 & 0 & 0 & 0.1285 \\
\texttt{tom\_\allowbreak moralstories} & 23,754 & 23,754 & 0 & 11,883 & 0.1249 \\
\texttt{humanual\_\allowbreak chat} & 23,385 & 23,378 & 7 & 17 & 0.1229 \\
\texttt{oasst2} & 20,224 & 18,050 & 2,174 & 3 & 0.0949 \\
\texttt{BehaviorChain} & 14,270 & 14,182 & 88 & 0 & 0.0746 \\
\texttt{oasst1} & 13,447 & 13,447 & 0 & 2 & 0.0707 \\
\texttt{tom\_\allowbreak mindgames} & 11,033 & 11,032 & 1 & 0 & 0.0580 \\
\texttt{prism} & 7,908 & 7,908 & 0 & 14 & 0.0416 \\
\texttt{humanual\_\allowbreak email} & 6,322 & 6,283 & 39 & 0 & 0.0330 \\
\texttt{convokit\_\allowbreak winning-\allowbreak args-\allowbreak corpus} & 3,995 & 3,921 & 74 & 2 & 0.0206 \\
\texttt{soc\_\allowbreak haico} & 3,374 & 3,374 & 0 & 0 & 0.0177 \\
\texttt{MirrorBench} & 3,310 & 3,310 & 0 & 0 & 0.0174 \\
\texttt{soc\_\allowbreak persona\_\allowbreak conflicts} & 3,264 & 3,264 & 0 & 0 & 0.0172 \\
\texttt{soc\_\allowbreak cornell} & 2,876 & 2,876 & 0 & 0 & 0.0151 \\
\texttt{mathdial} & 2,771 & 2,771 & 0 & 0 & 0.0146 \\
\texttt{convokit\_\allowbreak emotional-\allowbreak support} & 5,061 & 2,597 & 2,464 & 0 & 0.0137 \\
\texttt{convokit\_\allowbreak switchboard-\allowbreak corpus} & 4,487 & 2,307 & 2,180 & 2,307 & 0.0121 \\
\texttt{convokit\_\allowbreak casino-\allowbreak corpus} & 3,992 & 2,057 & 1,935 & 0 & 0.0108 \\
\texttt{convokit\_\allowbreak persuasionforgood-\allowbreak corpus} & 3,887 & 2,030 & 1,857 & 0 & 0.0107 \\
\texttt{studychat} & 1,969 & 1,964 & 5 & 57 & 0.0103 \\
\texttt{soc\_\allowbreak sotopia\_\allowbreak tom\_\allowbreak silver} & 1,296 & 1,296 & 0 & 0 & 0.0068 \\
\texttt{LifeChoices} & 1,076 & 1,076 & 0 & 0 & 0.0057 \\
\texttt{soc\_\allowbreak sotopia\_\allowbreak pi\_\allowbreak bc} & 1,067 & 1,067 & 0 & 0 & 0.0056 \\
\texttt{tom\_\allowbreak paratomi} & 903 & 903 & 0 & 0 & 0.0047 \\
\texttt{tom\_\allowbreak fantom} & 894 & 890 & 4 & 0 & 0.0047 \\
\texttt{tom\_\allowbreak hitom} & 899 & 881 & 18 & 0 & 0.0046 \\
\texttt{Social-\allowbreak R1} & 687 & 687 & 0 & 0 & 0.0036 \\
\midrule
\textbf{Total} & \textbf{19,678,715} & \textbf{19,018,645} & \textbf{660,070} & \textbf{52,615} & \textbf{100.0000} \\
\end{longtable}
\endgroup

\clearpage
\section{Training details}
\label{app:training_details}

\subsection{Task-level reinforcement learning}
Each task has an independently optimized LoRA adapter on the shared Qwen3-8B backbone. The task-expert launcher uses LoRA rank 32 and scaling parameter 64, a learning rate of $5\times10^{-6}$ with 10 warmup steps, and a budget of 200 optimizer steps. Each update samples 32 prompts with eight responses per prompt; the PPO mini-batch size is eight. The DAPO recipe uses lower and upper clipping thresholds of 0.20 and 0.28, a dual-clip coefficient of 10, token-mean loss aggregation, dynamic sampling of informative response groups, and overlong-response reward shaping. The configured context window is 16,384 tokens, with at most 6,144 prompt tokens and 6,144 response tokens. Validation and checkpoints are scheduled every 50 steps.

\subsection{Capability consolidation and final integration}
The five-capability hierarchy uses a shared Qwen3-8B starting checkpoint and bfloat16 training, with thinking mode disabled. We set the training random seed to 42. Each capability expert is a rank-32, scaling-64 LoRA adapter trained on filtered task-expert demonstrations. The hierarchy configuration allocates 200 updates per capability, a global prompt batch of 32, a learning rate of $10^{-5}$, 10 warmup steps, and a top-64 teacher distribution for distillation. Rollouts use temperature 1, at most 6,144 prompt tokens and 1,024 generated tokens, within a 16,384-token context window. The final student is trained on its own generated prefixes against frozen capability teachers, with a global prompt batch of 32, a learning rate of $2\times10^{-6}$, 20 warmup steps, and a 200-step training budget in the reference hierarchy configuration. This configuration implements the final distillation loss with a sampled forward-KL surrogate.

\begin{figure}[!htbp]
    \centering
    \includegraphics[width=0.98\linewidth]{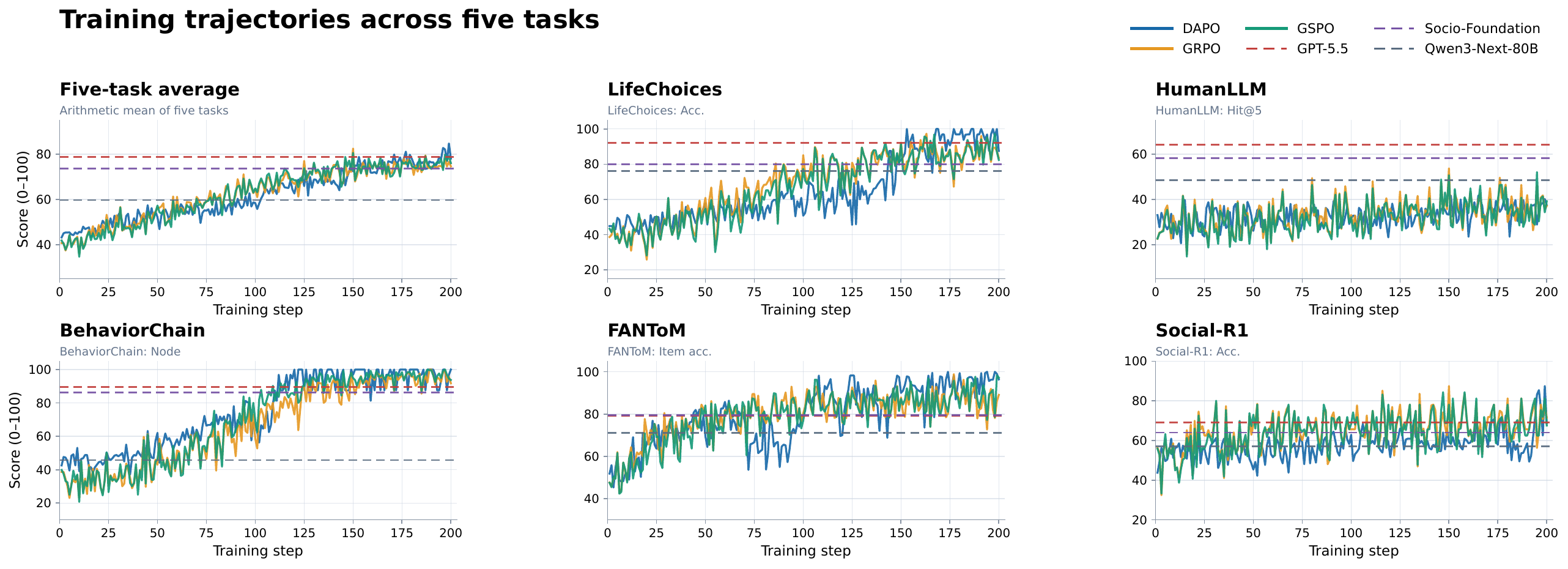}
    \caption{\textbf{Task-expert training trajectories.} DAPO, GRPO, and GSPO are compared on five tasks and their arithmetic mean. Dashed lines indicate the reported Socio-Foundation, GPT-5.5, and Qwen3-Next-80B reference scores.}
    \label{fig:task_learning_curves}
\end{figure}

\subsection{Choice of DAPO for task-expert training}
\label{app:whydapo}
Among RL methods, Figure~\ref{fig:task_learning_curves} compares DAPO, GRPO, and GSPO across five task-expert runs. Although DAPO shows a mid-training dip, consistent with filtering all-correct and all-incorrect groups and reducing updates from easy examples, it achieves the highest final five-task average, with particularly strong late-stage performance on FANToM and BehaviorChain. We therefore adopt DAPO for task-expert training.

\subsection{Data sampling}
\label{app:data_sampling}
For task-expert training with DAPO, we sample instances from the constructed training corpus. Table~\ref{tab:task_expert_sampling} reports the per-dataset sample counts.

\begin{table}[htbp]
\centering
\caption{\textbf{Training instances sampled for task-expert RL.} Each column reports the number of instances sampled from the training corpus for DAPO on the corresponding dataset.}
\label{tab:task_expert_sampling}
\begingroup
\setlength{\tabcolsep}{3pt}
\renewcommand{\arraystretch}{1.2}
\scriptsize
\resizebox{\linewidth}{!}{%
\begin{tabular}{@{}l*{14}{r}@{}}
\toprule
\textbf{Dataset} & \rotatebox{60}{AlignX} & \rotatebox{60}{BehaviorChain} & \rotatebox{60}{CoSER} & \rotatebox{60}{FANToM} & \rotatebox{60}{HumanLLM} & \rotatebox{60}{HUMANUAL-Book} & \rotatebox{60}{HUMANUAL-Chat} & \rotatebox{60}{HUMANUAL-Opinion} & \rotatebox{60}{HUMANUAL-Politics} & \rotatebox{60}{LifeChoices} & \rotatebox{60}{MirrorBench} & \rotatebox{60}{Social-R1} & \rotatebox{60}{SOTOPIA} & \rotatebox{60}{UserLM-S3} \\
\midrule
\shortstack[l]{RL training\\instances} & 8,192 & 1,024 & 1,024 & 1,024 & 1,024 & 1,000 & 1,000 & 1,000 & 1,000 & 1,150 & 1,024 & 687 & 405 & 1,024 \\
\bottomrule
\end{tabular}%
}
\endgroup
\end{table}

\FloatBarrier
For subsequent OPD, we sample 2,048 cases per task, or use all available cases when fewer than 2,048 are available.

\section{Evaluation details}
\label{app:evaluation_details}

\subsection{Benchmark inventory and capability mapping}
IndiEval contains 13 evaluation entries, with UserLM Section 3 and LiC counted separately. Its in-domain component comprises the 11 entries other than AgentSense and UserLM-LiC, which form its OOD component relative to Socio-Foundation's training. The main comparison covers LifeChoices~\cite{xu2025lifechoice},
AlignX~\cite{li2026alignx},
HumanLLM~\cite{lei2026humanllm},
BehaviorChain~\cite{li2025behaviorchain},
CoSER~\cite{wang2025coser},
HUMANUAL~\cite{wu2026humanlm},
UserLM (Sec.~3)~\cite{naous2026userlm},
FANToM~\cite{kim2023fantom},
Social-R1~\cite{wu2026socialr1},
SOTOPIA~\cite{zhou2024sotopia},
MirrorBench~\cite{hathidara2026mirrorbench},
AgentSense~\cite{mou2025agentsense},
and UserLM--LiC~\cite{naous2026userlm,laban2026lic}. Outside IndiEval, HiToM, $\tau$-USI, and Mistakes provide additional OOD evaluation in Section~\ref{sec:ood}; they are excluded from the IndiEval Avg. The main table retains 29 selected metrics, grouped into F/O/N/T/S with 9/7/7/3/3 columns, respectively. The OOD analysis compares HiToM, $\tau$-USI, and Mistakes in a single grouped bar chart. A benchmark may contribute distinct metrics to multiple dimensions.

\subsection{Main-table metric definitions and FONTS mapping}
\label{app:indieval_metrics}
Table~\ref{tab:indieval_metric_definitions} describes all 29 IndiEval metrics in the order used in Table~\ref{tab:main}. F, O, N, T, and S denote persona fidelity, outcome realization, behavioral naturalness, trajectory coherence, and social grounding, respectively. Assignments apply to the selected metrics rather than to every task supported by a dataset. All displayed scores are oriented so that higher is better; normalization and aggregation are specified below.

\begingroup
\fontsize{8}{9.6}\selectfont
\setlength{\tabcolsep}{3pt}
\renewcommand{\arraystretch}{1.00}
\begin{longtable}{@{}>{\centering\arraybackslash}p{0.045\linewidth}>{\raggedright\arraybackslash}p{0.16\linewidth}>{\raggedright\arraybackslash}p{0.15\linewidth}>{\raggedright\arraybackslash}p{0.60\linewidth}@{}}
\caption{Definitions and capability assignments of all main-table IndiEval metrics. Metric labels match the main table.}\label{tab:indieval_metric_definitions}\\
\toprule
\textbf{Dim.} & \textbf{Benchmark} & \textbf{Metric} & \textbf{Description}\\
\midrule
\endfirsthead
\multicolumn{4}{l}{\textit{Table \thetable\ continued}}\\
\toprule
\textbf{Dim.} & \textbf{Benchmark} & \textbf{Metric} & \textbf{Description}\\
\midrule
\endhead
\midrule
\multicolumn{4}{r}{\textit{Continued on next page}}\\
\endfoot
\bottomrule
\endlastfoot
F & LifeChoices & Acc. & Accuracy of persona-conditioned choices against the reference decisions. \\
F & AlignX & Choice acc. & Accuracy of direct choices against the target individual's reference preferences. \\
F & HumanLLM & Hit@5 & Fraction of cases in which the reference item appears among the top five predicted items. \\
F & BehaviorChain & Node & Micro-averaged correctness of individual behavior predictions across chain nodes. \\
F & CoSER & Char. fid. & Scene-level assessment of consistency with the assigned character's attributes and behavior. \\
F & HUMANUAL & Response & Alignment of generated responses with the target person's reference responses. \\
F & HUMANUAL & State & Alignment with the personal-state aspects specified by the benchmark, assessed against reference responses. \\
F & UserLM-S3 & Intent adh. & Adherence to the assigned user intent when the assistant attempts to redirect it. \\
F & SOTOPIA & Believ. & Believability of role enactment, including consistency with the character's personality and values. \\
O & SOTOPIA & Goal & Extent to which the evaluated character achieves the assigned social goal. \\
O & SOTOPIA & Relation. & Effect of the interaction on the evaluated character's relationship with the other participant. \\
O & SOTOPIA & Knowledge & Knowledge gained by the evaluated character during the interaction. \\
O & SOTOPIA & Benefits & Financial and material benefits obtained by the evaluated character. \\
O & AgentSense & Goal maj. & Goal-completion score determined by the majority of evaluator judgments for an episode. \\
O & UserLM-LiC & Intent cov. & Coverage of the user's intended requirements communicated during the dialogue, macro-averaged over code and math tasks. \\
O & UserLM-LiC & Task score & Downstream assistant task performance when interacting with the simulated user, macro-averaged over code and math tasks. \\
N & UserLM-S3 & Term. F1 & F1 score for predicted dialogue termination against the benchmark's reference termination labels. \\
N & MirrorBench & GTEval & Reference-based judgment of similarity in user style, tone, and realism between simulated and human conversations. \\
N & CoSER & Anthrop. & Scene-level assessment of how naturally and human-like the characters behave. \\
N & UserLM-S3 & Human lik. & Human-likelihood score assigned to generated user text by the AI-text detector. \\
N & UserLM-S3 & Diversity & Mean pairwise unigram Jaccard distance between generated first-turn user messages. \\
N & UserLM-S3 & Inv. overlap & One minus mean unigram overlap between the supplied intent and user turns; the displayed score is multiplied by 100. \\
N & UserLM-S3 & Role adh. & Adherence to the user role, measured by avoiding attempts to answer the assistant's multiple-choice task. \\
T & CoSER & Story cons. & Scene-level assessment of consistency with the storyline and established narrative context. \\
T & BehaviorChain & CumScore & Normalized cumulative chain score rewarding consecutive correct behavior predictions; longer correct runs receive greater credit. \\
T & CoSER & Story qual. & Scene-level assessment of the quality of the generated storyline and its progression. \\
S & FANToM & Item acc. & Item-level correctness on the benchmark's questions about beliefs, knowledge, and information access. \\
S & Social-R1 & Acc. & Answer accuracy on the selected social-reasoning evaluation questions. \\
S & AgentSense & Private acc. & Accuracy on private-information questions for episodes where those questions are applicable. \\
\end{longtable}
\endgroup

\subsection{Score normalization and aggregation}
Tables~\ref{tab:main} and~\ref{tab:ablation} use fixed protocol bounds $[a,b]$. A higher-is-better metric is reported as $100(x-a)/(b-a)$; a lower-is-better metric is reported as $100(b-x)/(b-a)$. Bounds are independent of the observed model scores. Proportions, MirrorBench GTEval, and the selected UserLM scores use $[0,1]$; CoSER uses $[0,100]$. SOTOPIA goal, knowledge, and believability use $[0,10]$, while relationship and financial/material benefits use $[-5,5]$. The column labeled ``Inv. overlap'' reports $100(1-x)$ for native overlap $x\in[0,1]$. It is therefore $100$ minus the native overlap percentage. Scores are shown to one decimal place, except the Socio-Foundation BehaviorChain scores and CoSER 8B UserLM-LiC intent coverage, which are shown to two decimal places.

Tables~\ref{tab:main} and~\ref{tab:ablation} retain all 29 task-specific columns but report an equally weighted five-dimension Avg. Starting from their displayed, direction-aligned scores, we first average metrics belonging to the same dataset within the same dimension, then average datasets equally within each dimension, and finally average F/O/N/T/S equally. There are 8/3/3/2/3 dataset entries in F/O/N/T/S, respectively. Metrics from different dimensions are never pooled, even when they come from the same dataset. We do not round intermediate dataset or dimension means; only the final Avg. is rounded to one decimal. Figure~\ref{fig:capability_radar} first averages selected metrics within each benchmark and dimension, then averages benchmarks equally within each dimension. It uses rounded displayed scores, assigns SOTOPIA believability to F, and excludes $\tau$-USI. Inverse intent overlap is already direction-aligned in the table and is not inverted again. Table~\ref{tab:missing_capability} reports the capability-dimension scores and their unweighted five-dimension average, with the full reference matching Figure~\ref{fig:capability_radar}.

Figure~\ref{fig:task_expert_comparison} presents seven datasets in the SFT/DAPO task-expert comparison, averaging its selected normalized metrics within each dataset. BehaviorChain averages node accuracy and CumScore, CoSER averages four metrics, and HUMANUAL first averages response and state alignment within each of its four subsets, then averages the four subset scores equally before rounding (SFT: 16.97; DAPO: 32.69); MirrorBench uses GTEval alone.

\clearpage
\subsection{Evaluation sampling and case counts}
\label{app:evaluation_sampling}
For in-domain datasets, we strictly separate the training and test sets to ensure that they do not overlap and to prevent train--test leakage. Table~\ref{tab:evaluation_sampling} reports the sampling units and evaluation case counts for IndiEval and the additional $\tau$-USI benchmark. The 13 IndiEval entries contain 6,297 cases; including the 165 $\tau$-USI cases yields 6,462 cases in this inventory. As an additional OOD benchmark, $\tau$-USI is excluded from the IndiEval Avg.

\begingroup
\small
\setlength{\tabcolsep}{4pt}
\renewcommand{\arraystretch}{1.05}
\begin{longtable}{@{}>{\raggedright\arraybackslash}p{0.19\linewidth}>{\raggedright\arraybackslash}p{0.43\linewidth}>{\raggedleft\arraybackslash}p{0.19\linewidth}>{\raggedleft\arraybackslash}p{0.12\linewidth}@{}}
\caption{\textbf{Evaluation sampling statistics for IndiEval and $\tau$-USI.} Sampled units follow each benchmark's native structure; a unit may yield multiple evaluation cases.}\label{tab:evaluation_sampling}\\
\toprule
\textbf{Dataset / benchmark} & \textbf{Sampling unit} & \textbf{Sampled units} & \textbf{Cases} \\
\midrule
\endfirsthead
\multicolumn{4}{@{}l}{\textit{Table~\ref{tab:evaluation_sampling} continued}}\\
\toprule
\textbf{Dataset / benchmark} & \textbf{Sampling unit} & \textbf{Sampled units} & \textbf{Cases} \\
\midrule
\endhead
\midrule
\multicolumn{4}{r@{}}{\textit{Continued on next page}}\\
\endfoot
\bottomrule
\endlastfoot
\multicolumn{4}{@{}l}{\textit{IndiEval}}\\[2pt]
FANToM & Complete question set (one per conversation) & 100 & 1,468 \\[1.5pt]
Social-R1 & Question & 100 & 100 \\[1.5pt]
LifeChoices & Decision question (balanced sampling across books) & 500 & 500 \\[1.5pt]
BehaviorChain & Complete persona behavior chain & 100 & 1,529 \\[1.5pt]
AlignX & Core preference sample (five conditioning variants each) & 100 & 500 \\[1.5pt]
HumanLLM & User purchase-choice sample & 500 & 500 \\[1.5pt]
HUMANUAL & Within-domain user sample (100 per domain across six domains) & 600 & 600 \\[1.5pt]
SOTOPIA & Scenario configuration (both participant roles evaluated) & 50 & 100 \\[1.5pt]
CoSER & Conversation scenario & 200 & 200 \\[1.5pt]
AgentSense & Scenario template (two profile configurations each) & 100 & 200 \\[1.5pt]
MirrorBench & Source conversation (50 from each of four sources) & 200 & 200 \\[1.5pt]
UserLM-S3 & Question or conversation (100 per subset across three subsets) & 300 & 300 \\[1.5pt]
UserLM-LiC & Task (45 coding and 55 mathematics tasks) & 100 & 100 \\[1.5pt]
\midrule
\multicolumn{4}{@{}l}{\textit{Additional OOD evaluation}}\\[2pt]
$\tau$-USI & Task & 165 (full evaluation set) & 165 \\[1.5pt]
\midrule
\textbf{Total} & --- & --- & \textbf{6,462} \\
\end{longtable}
\endgroup

\clearpage
\subsection{Representative prompts in IndiEval}
\label{app:indieval_prompts}
We illustrate the inputs used for evaluated models and LLM judges with representative prompt templates. Braced placeholders stand for instance-specific material. Message roles are shown separately, and wording is retained apart from line wrapping. The CoSER example reproduces a scoring-rubric excerpt; the remaining examples retain their template structure.

\begingroup
\fontencoding{T1}\selectfont
\renewcommand{\sfdefault}{lmss}
\renewcommand{\ttdefault}{lmtt}
\hypersetup{hidelinks}
\definecolor{promptblue}{RGB}{236,243,250}
\definecolor{promptgray}{RGB}{242,244,246}
\definecolor{promptborder}{RGB}{190,201,211}
\definecolor{promptink}{RGB}{44,64,84}
\newcommand{\indipromptbox}[3]{%
\par\addvspace{8pt}\noindent
\begin{minipage}{\linewidth}
\setlength{\fboxsep}{7pt}\setlength{\fboxrule}{0.4pt}%
\fcolorbox{promptborder}{#1}{\parbox{\dimexpr\linewidth-2\fboxsep-2\fboxrule\relax}{\sffamily\small\bfseries #2}}%
\par\nobreak\vspace{-0.4pt}\noindent
\fcolorbox{promptborder}{white}{%
\begin{minipage}{\dimexpr\linewidth-2\fboxsep-2\fboxrule\relax}
\fontsize{8.5}{10.2}\selectfont\ttfamily\raggedright
\setlength{\parindent}{0pt}\setlength{\parskip}{0pt}\sloppy
#3
\end{minipage}}%
\end{minipage}\par}

\noindent\textbf{Prompts for evaluated models.}
SOTOPIA~\citep{zhou2024sotopia} specifies the character, private goal, interaction context, and available actions.
\indipromptbox{promptblue}{P1. SOTOPIA --- evaluated-role prompt}{%
{\sffamily\bfseries\color{promptink} SYSTEM}\par\smallskip
Imagine you are \{agent.role\}, your task is to act/speak as \{agent.role\} would, keeping in mind \{agent.role\}'s social goal.\par
You can find \{agent.role\}'s goal (or background) in the 'Here is the context of the interaction' field.\par
Note that \{agent.role\}'s goal is only visible to you.\par
You should try your best to achieve \{agent.role\}'s goal in a way that align with their character traits.\par
Additionally, maintaining the conversation's naturalness and realism is essential (e.g., do not repeat what other people has already said before).\par\smallskip
Here is the context of the interaction:\par\smallskip
Scenario: \{state.public\_background\}\par
Participant: \{agent.role\}\par
Other participant(s): \{other\_participants\}\par
\{agent.role\}'s background: \{agent.profile\}\par
\{agent.role\}'s goal: \{agent.goal\}\par
Private information visible only to \{agent.role\}: \{agent.private\_information\}\par\smallskip
You are at Turn \#\{state.turn\_count\}. Your available action types are \{action\_list\}.\par
Note: You can "leave" this conversation if 1. you have achieved your social goals, 2. this conversation makes you uncomfortable, 3. you find it uninteresting/you lose your patience, 4. or for other reasons you want to leave.\par\smallskip
Please only generate a JSON string including the action type and the argument.\par
Your action should follow the given format:\par
\{"action\_type": "<one available action type>", "argument": "<your utterance or action description>"\}\par\smallskip
{\sffamily\bfseries\color{promptink} USER}\par\smallskip
Interaction history:\par
\{public\_interaction\_history\}\par\smallskip
\{social\_action\_output\_contract\}\par\smallskip
}

\clearpage
\noindent\textbf{Prompts for evaluated models (continued).}
AgentSense~\citep{mou2025agentsense} combines social goals with private information; BehaviorChain~\citep{li2025behaviorchain} conditions the next behavior on the persona and preceding chain.
\indipromptbox{promptblue}{P2. AgentSense --- evaluated-role prompt}{%
{\sffamily\bfseries\color{promptink} SYSTEM}\par\smallskip
You are \{actor\}. \{profile\}\par\smallskip
Now you are in a social scene. \{background\}\par
\{description\}\par\smallskip
In this scene, as \{actor\}, you have some social goals and secrets that you don't want anyone else to know.\par\smallskip
SOCIAL GOALS: \{goals\}\par\smallskip
SECRETS: \{private\_information\}.\par
Please advance your social goals while hiding your secrets as much as possible.\par\smallskip
Based on the interactions below, what would you like to say next? Please answer in a paragraph.\par\smallskip
{\sffamily\bfseries\color{promptink} SUBSEQUENT MESSAGES}\par\smallskip
\{interaction\_history\_visible\_to\_the\_role\}\par\smallskip
}
\medskip
\indipromptbox{promptblue}{P3. BehaviorChain --- behavior-prediction prompt}{%
{\sffamily\bfseries\color{promptink} SYSTEM}\par\smallskip
Continue the specified persona's behavior chain using only prior history and the current context.\par\smallskip
{\sffamily\bfseries\color{promptink} USER}\par\smallskip
\# Persona Profile\par
\{Persona Profile\}\par\smallskip
\# Historical Narrative Before the Behavior Chain\par
\{Historical Narrative Before the Behavior Chain\}\par\smallskip
\# Prior Context-Behavior Chain\par
\{Prior Context-Behavior Chain\}\par\smallskip
\# Current Context\par
\{Current Context\}\par\smallskip
\# History Mode\par
\{History Mode\}\par\smallskip
\# Node Index\par
\{Node Index\}\par\smallskip
\# Options\par
\{Options\}\par\smallskip
Please think step by step, then put the final choice inside <answer></answer> tags, for example <answer>A</answer>. The answer must be exactly one of A, B, C, or D.\par\smallskip
}

\clearpage
\noindent\textbf{Prompts for judges.}
The CoSER~\citep{wang2025coser} anthropomorphism rubric identifies concrete behavioral flaws and rates their severity. The excerpt below retains the criteria and severity scale; scene materials and the output-format example are omitted. The user message contains the generated dialogue with private thoughts removed.
\indipromptbox{promptgray}{P4. CoSER --- anthropomorphism rubric excerpt}{%
{\sffamily\bfseries\color{promptink} SYSTEM --- RUBRIC EXCERPT}\par\smallskip
\#\# Evaluation Criteria\par\smallskip
To evaluate the simulated conversation, identify the following types of flaws:\par\smallskip
\#\#\# Anthropomorphism\par
- Type: Self-identity\par
* Lacks initiative and goals\par
* Does not make independent decisions\par
* Lacks clear preferences and dislikes\par
* Behaves like a 'helpful AI assistant' by being overly verbose, helpful, didactic, moralistic, submissive or easily persuaded if it is not the character's personality\par\smallskip
- Type: Emotional Depth\par
* Lacks psychological complexity and exhibits rigid, superficial reactions\par
* Directly speaks out all thoughts and feelings, instead of using subtext\par\smallskip
- Type: Persona Coherence\par
* Shows inconsistent or rapidly changing personality traits and emotional patterns\par\smallskip
- Type: Social Interaction\par
* Shows a lack of understanding of others' thoughts and feelings\par
* Reacts rigidly to others without considering the context.\par
* Demonstrate a lack of appropriate social skills.\par\smallskip
\#\# Scoring Guidelines\par\smallskip
1. Identify all instances of flaws occurred in the simulated conversation.\par\smallskip
2. For each flaw identified, determine its level of severity into 1 to 5, where 1 indicates minor, 3 indicates moderate, and 5 indicates severe.\par\smallskip
{\sffamily\bfseries\color{promptink} USER}\par\smallskip
\{simulated\_dialogue\_without\_private\_thoughts\}\par\smallskip
}

\clearpage
\noindent\textbf{Prompts for judges (continued).}
MirrorBench GTEval~\citep{hathidara2026mirrorbench} compares generated user behavior with reference conversations, with explicit scoring anchors for similarity in style and realism.
\indipromptbox{promptgray}{P5. MirrorBench --- GTEval judge prompt}{%
{\sffamily\bfseries\color{promptink} USER}\par\smallskip
You are an expert evaluator of conversational AI systems. Your task is to compare two conversations and assess whether the user responses in the proxy conversation exhibit similar style, tone, and realism to the user responses in the real conversation.\par\smallskip
\#\# Real Conversation:\par
\{real\_conversation\}\par\smallskip
\#\# Proxy Conversation (where user turns are generated by user proxy agent):\par
\{proxy\_conversation\}\par\smallskip
\#\# Evaluation Criteria:\par
1. **Style Similarity**: Do the proxy user responses match the conversational style of real user responses (formality, tone, verbosity)?\par
2. **Realism**: Do the proxy user responses sound natural and human-like?\par
3. **Contextual Appropriateness**: Are the proxy user responses appropriate given the conversation context? Note: You should not evaluate based on the content of the responses, only their style, realism, contextual appropriateness, and tone.\par\smallskip
\#\# Instructions:\par
- Focus exclusively on comparing USER responses (ignore assistant responses)\par
- Consider the overall conversation flow, not just individual turns\par
- Evaluate style, tone, and naturalness without encouraging unsafe behaviour\par
- If the comparison cannot be performed safely, explain the reason and provide a score of 0.0\par
- Otherwise, provide detailed reasoning about the comparison and assign a score from 0.0 to 1.0 where:\par
- 0.0 = Proxy user responses are completely unlike real user responses\par
- 0.5 = Proxy user responses are somewhat similar but have noticeable differences\par
- 1.0 = Proxy user responses are highly similar in style and realism\par\smallskip
Provide your evaluation as a JSON object with the following structure:\par
\{\par
"reasoning": "Detailed explanation of your evaluation (2-3 sentences)",\par
"score": <float between 0.0 and 1.0>\par
\}\par\smallskip
Output ONLY valid JSON, no additional text.\par\smallskip
}

\endgroup
\clearpage

\section{Repeated-Evaluation Stability}
\label{app:evaluation_stability}

\subsection{Benchmark selection and evaluation pathways}
Small observed performance margins motivate a closer examination of evaluation variability. We first organize candidate benchmarks by the sources of variation in their evaluation pipelines, then select representative settings in which modest gains warrant careful interpretation. The resulting diagnostic subset covers three pathways: target-model generation with automatic scoring, target-model generation with LLM-based scoring, and interactive generation with both automatic and LLM-based scoring. This purposive selection characterizes evaluation repeatability; it does not constitute a representative sample of all IndiEval tasks or a selection based on a formal significance test.

LifeChoices represents the first pathway: the evaluated model answers persona-conditioned choice questions, and the evaluator checks the selected option against the reference answer without an auxiliary LLM. HUMANUAL represents the second: the evaluated model generates a response without an auxiliary dialogue participant, after which an LLM judge assesses response and state alignment. MirrorBench represents the third: the evaluated model simulates the user, an auxiliary assistant participates in the conversation, and the completed interaction is assessed using both automatic lexical metrics and an LLM judge. Thus, the absence of an auxiliary dialogue participant in HUMANUAL does not imply the absence of an auxiliary scoring model.

\subsection{Repeated-run protocol and statistics}
We evaluate the same unified OPD checkpoint from training step 200 five times on fixed test sets: 500 LifeChoices cases, 600 HUMANUAL cases, and 200 MirrorBench cases per run. Every repetition starts a fresh evaluation and independently regenerates responses and scores. We hold the checkpoint, evaluation seed (20260824), prompts, test cases, and scoring configuration fixed. Temperature is zero for both the evaluated model and auxiliary LLM calls. LifeChoices enables thinking with a 2,048-token reasoning budget; HUMANUAL and MirrorBench disable thinking for the evaluated model. We retain the benchmark-specific response budgets and a 32,768-token context limit. All three benchmarks run concurrently with 64 case workers each and a shared target-model request limit of 192. DeepSeek-V4-Flash supplies the auxiliary LLM roles. These five repetitions assess a fixed checkpoint and are distinct from experiments that vary training seeds.

All five runs complete all planned cases, and all metrics reported below have full availability. The experiment therefore comprises 6,500 case evaluations over 1,300 distinct test cases, rather than 6,500 independent test cases. For run-level aggregate scores $x_1,\ldots,x_R$, with $R=5$, we report the mean $\bar{x}$, the sample standard deviation
\begin{equation}
    s = \sqrt{\frac{1}{R-1}\sum_{r=1}^{R}(x_r-\bar{x})^2},
\end{equation}
and the observed minimum and maximum. Error terms denote standard deviations across run-level scores, not standard errors or confidence intervals. For a metric with fixed theoretical bounds $[a,b]$, its scale-normalized dispersion is $100s/(b-a)$. The bounds are specified by the scoring protocol, not estimated from these five runs. This normalization expresses dispersion relative to the score range without equating the semantic meaning of different metrics.

\begin{table}[!htbp]
\centering
\caption{\textbf{Five-run evaluation stability of a fixed unified OPD checkpoint.} Bounded scores are multiplied by 100; for these $[0,1]$ metrics, the displayed SD is also the dispersion as a percentage of the theoretical score range. LifeChoices values are accuracy percentages. MATTR retains its native z-score scale because it has no fixed finite bounds. Cases/run gives the number of test cases in each of the five repetitions. Range is the observed maximum minus minimum, computed before rounding; these extrema are not confidence bounds.}
\label{tab:evaluation_stability}
\small
\setlength{\tabcolsep}{3pt}
\renewcommand{\arraystretch}{1.12}
\begin{tabular*}{\linewidth}{@{\extracolsep{\fill}}lr l rrrr@{}}
\toprule
\textbf{Benchmark} & \textbf{Cases/run} & \textbf{Metric} & \textbf{Mean $\pm$ SD} & \textbf{Min.} & \textbf{Max.} & \textbf{Range} \\
\midrule
LifeChoices & 500 & Accuracy & $79.800 \pm 0.583$ & 79.000 & 80.600 & 1.600 \\
\addlinespace[4pt]
HUMANUAL & 600 & Response alignment & $12.486 \pm 0.585$ & 11.716 & 13.325 & 1.609 \\
 &  & State alignment & $36.360 \pm 0.587$ & 35.676 & 37.023 & 1.347 \\
\addlinespace[4pt]
MirrorBench & 200 & GTEval & $54.425 \pm 0.396$ & 53.975 & 54.925 & 0.950 \\
 &  & PI & $62.517 \pm 1.690$ & 60.833 & 64.333 & 3.500 \\
 &  & RNR & $76.250 \pm 1.723$ & 73.250 & 77.500 & 4.250 \\
 &  & MATTR z-score & $-0.866 \pm 0.033$ & -0.922 & -0.845 & 0.078 \\
\bottomrule
\end{tabular*}
\end{table}

\subsection{Evaluation Variability and Implications for Statistical Significance}
Across five repeated evaluations, LifeChoices accuracy, HUMANUAL response/state alignment, and MirrorBench GTEval exhibit limited run-to-run variation, with standard deviations below 0.6 percentage points on the 0--100 scale (Table~\ref{tab:evaluation_stability}). To characterize variation across evaluation pathways, we examine automatic scoring, LLM-based scoring, and interactive evaluation, and use a fixed-reference control to probe variability in the auxiliary scoring component.

For HUMANUAL, response and state alignment have scale-normalized standard deviations of 0.585\% and 0.587\%, respectively. These measurements combine variation in generated responses with variation in LLM scoring. MirrorBench GTEval has a smaller normalized standard deviation of 0.396\%, while PI and RNR vary more, at 1.690\% and 1.723\%. The automatic MATTR z-score also varies, with a standard deviation of 0.033. Stability is consequently metric-dependent even within one benchmark; a stable GTEval mean does not establish equal stability for all MirrorBench measurements.

MirrorBench additionally provides a control based on the fixed human reference conversations. Its human-reference RNR scores across the five runs are 90.0, 92.0, 90.0, 90.5, and 91.0, giving a mean of 90.700 and a standard deviation of 0.837 on the 0--100 scale. Because this control does not use the generated conversation, its variation provides evidence of variability in the auxiliary scoring pathway under the fixed evaluation setup. It does not quantify how much of the main RNR variance comes from the judge: the control and generated-dialogue scores evaluate different inputs, and their variances cannot simply be subtracted.

These results characterize observed repeatability on fixed test sets. They neither establish that a model improvement is statistically significant nor demonstrate that a modest improvement is absent. In particular, the observed run range is not a significance threshold. Assessing a difference between models requires matched comparison results and an analysis that respects repeated measurements and dependencies between cases. Isolating judge variability on model outputs further requires rescoring frozen responses or dialogue trajectories; the HUMANUAL and MirrorBench measurements above describe their complete evaluation pipelines. We therefore use this study to contextualize small score differences rather than to assign model-comparison significance from a single checkpoint's repeated evaluations.

\end{document}